\documentclass[10pt,twocolumn,letterpaper]{article}

\usepackage[pagenumbers]{cvpr} 
\usepackage{tikz}
\usepackage{pifont}
\usepackage[table]{xcolor}  

\usepackage{multirow}
\definecolor{cvprblue}{rgb}{0.21,0.49,0.74}
\usepackage[pagebackref,breaklinks,colorlinks,allcolors=cvprblue]{hyperref}

\def\paperID{15324} 
\def\confName{CVPR}
\def\confYear{2026}

\title{WiseEdit: Benchmarking Cognition- and Creativity-Informed Image Editing}

\author{Kaihang Pan$^{1}$\footnotemark[1] \quad Weile Chen$^{1}$\footnotemark[1]\quad  Haiyi Qiu$^{1}$\footnotemark[1] \quad Qifan Yu$^1$ \quad Wendong Bu$^{1}$ \quad Zehan Wang$^1$ \\
Yun Zhu$^{2}$\quad\quad Juncheng Li$^1$\quad\quad Siliang Tang$^{1}$\\ \small $^1$Zhejiang University, $^2$ Shanghai Artificial Intelligence Laboratory \\
{\tt\small \{kaihangpan, junchengli, siliang\}@zju.edu.cn}\\ \tt\normalsize Project Page: \url{https://qnancy.github.io/wiseedit_project_page/ } \\ \tt\normalsize \raisebox{-1.5pt}{\includegraphics[height=1.05em]{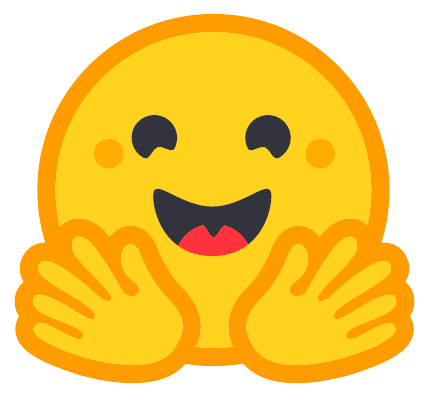}} Benchmark: \url{https://huggingface.co/datasets/123123chen/WiseEdit-Benchmark} \\ \tt\normalsize \raisebox{-1.5pt}{\includegraphics[height=1.05em]{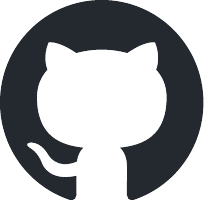}} Github: \url{https://github.com/beepkh/WiseEdit}
}

\begin{document}
\maketitle

\renewcommand{\thefootnote}{\fnsymbol{footnote}} 
\footnotetext[1]{Equal Contribution.} 

\begin{abstract}

Recent image editing models boast next-level intelligent capabilities, facilitating cognition- and creativity-informed image editing. Yet, existing benchmarks provide too narrow a scope for evaluation, failing to holistically assess these advanced abilities. To address this, we introduce WiseEdit, a knowledge-intensive benchmark for comprehensive evaluation of cognition- and creativity-informed image editing, featuring deep task depth and broad knowledge breadth. Drawing an analogy to human cognitive creation, WiseEdit decomposes image editing into three cascaded steps—Awareness, Interpretation, and Imagination—each corresponding to a task that poses a challenge for models to complete at the specific step. It also encompasses complex tasks, where none of the three steps can be finished easily. Furthermore, WiseEdit incorporates three fundamental types of knowledge: Declarative, Procedural, and Metacognitive knowledge. Ultimately, WiseEdit comprises 1,220 test cases, objectively revealing the limitations of SoTA image editing models in knowledge-based cognitive reasoning and creative composition capabilities. The benchmark, evaluation code, and the generated images of each model have already been publicly available. Project Page:  \url{https://qnancy.github.io/wiseedit_project_page/}.
\end{abstract}    

\section{Introduction}
\label{sec:intro}

\begin{figure*}[t]
\includegraphics[width=\linewidth]{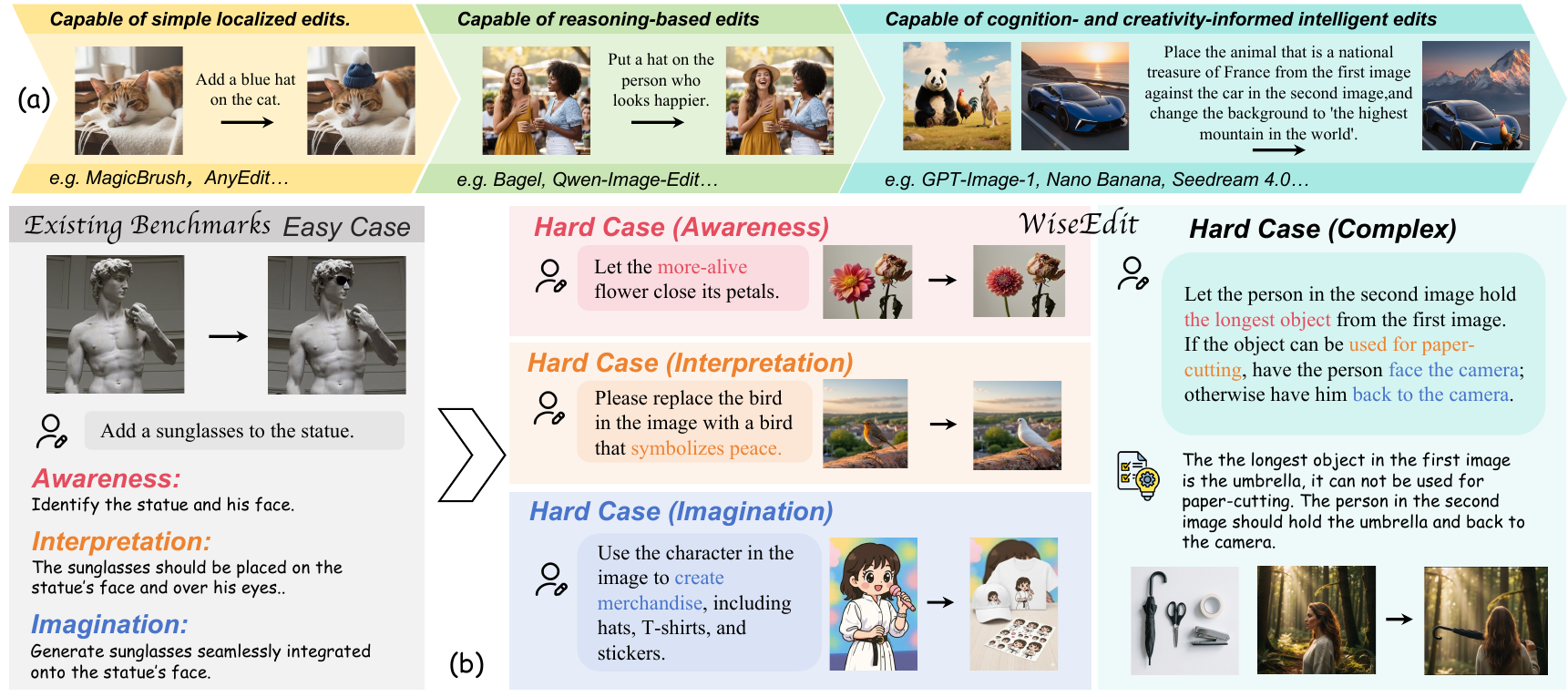}
\vspace{-1.8em}
\centering\caption{(a) Capabilities of different image editing models across varying skill levels. (b) Comparison between hard cases from WiseEdit and easy cases from existing simplistic image editing benchmarks.}
\label{fig:intro}
\vspace{-1.7em}
\end{figure*}

Instruction-based image editing~\cite{brooks2023instructpix2pix,zhao2024ultraedit} has attracted significant attention in recent years, offering general users a natural  way for image modification.
Early models~\cite{zhang2023magicbrush,yu2025anyedit} primarily focused on simple localized edits, such as object addition or removal.
Recently, remarkable progress in both visual understanding and visual generation, driven by multimodal large language models (MLLMs)~\cite{achiam2023gpt} and large diffusion models~\cite{esser2024scaling}, has fundamentally evolved this field.
As illustrated in Fig.~\ref{fig:intro} (a), models such as Bagel~\cite{deng2025emerging} and Qwen-Image-Edit~\cite{wu2025qwen} unify visual comprehension and generation in a single model to enable reasoning-based image editing.
Moreover, closed-source models like GPT-image-1~\cite{openai2024gpt4ocard} and Nano Banana~\cite{gemini2_5_flash_image, nanobananapro} further push the boundary toward cognition- and creativity-informed image editing. 
By simulating human cognition and creativity, these advanced systems facilitate semantic reasoning and multi-concept composition in complex editing scenarios.

To ensure these advanced models meet user practical needs, a comprehensive evaluation for  cognition- and creativity-informed image editing is imperative. 
Drawing an analogy with human cognition and creation~\cite{sweller2009cognitive}, we fundamentally decompose the image editing workflow into three interconnected step:
\textbf{(1) Awareness}, where the model establishes selective visual attention to precisely locate the modification target; 
\textbf{(2) Interpretation}, the cognitive core that parses editing instructions into directly executable, perception-level visual changes;
\textbf{(3) Imagination}, the generative engine that executes the creative realization to render a high-fidelity image.
Therefore, a robust evaluation must cover the \textbf{full task depth} of these three steps, verifying the capacity for knowledge-based reasoning and combinatorial creation, akin to the human faculty for converting abstract concepts into visual reality.
Moreover, to cover the \textbf{breadth of cognition}, models must demonstrate proficiency across three fundamental human knowledge~\cite{krathwohl2002revision}, declarative (knowing what), procedural (knowing how), and metacognitive (knowing about knowing), essential for robust problem-solving in intelligent image editing.

However, existing benchmarks fail to meet the above desiderata for cognition- and creativity-informed image editing.
Most~\cite{ju2023direct, zhang2023magicbrush, liu2025step1x} are too simplistic in their format and difficulty. They typically adopt a single input template as shown in Fig.~\ref{fig:intro}(b): one image paired with an explicit editing instruction that directly specifies how to modify concrete objects, which requires minimal cognitive or creative ability.
While newer benchmarks, such as KrisBench~\cite{wu2025kris}, have recently proposed to benchmark reasoning-informed image editing with world knowledge, they still fall short in task depth and cognitive breadth.
Specifically, they focus predominantly on knowledge-based reasoning during the interpretation phase, failing to cover the full spectrum of the three editing steps and also overlooking the assessment of meta-cognitive knowledge.
Consequently, the community urgently needs a holistic benchmark to more thoroughly measure the capabilities of current advanced models.

To address this gap, we introduce \textbf{WiseEdit}, a novel benchmark designed for cognition- and creativity-informed image editing tasks with rich task depth and knowledge breadth. 
As for task depth, WiseEdit features four types of challenging tasks across all three editing steps: \textbf{ Awareness Task}, \textbf{Interpretation Task}, \textbf{Imagination Task}, \textbf{WiseEdit-Complex}.
As shown in Fig.~\ref{fig:intro}(b).right, the first three tasks increase the difficulty specifically for their corresponding editing step, preventing trivial completion and requiring models to utilize causal reasoning or artistic creativity capabilities effectively.
The difficulty is further escalated by WiseEdit-Complex, where none of the three editing steps can be easily finished.
To further align with practical user needs, the task input is free-form, and each test case may include multiple input images.

Regarding knowledge breadth, we structure our tasks around three core types of knowledge to emulate human cognitive learning: \textbf{Declarative},\textbf{ Procedural}, and \textbf{Metacognitive knowledge}. 
Building upon this, we further broaden knowledge scope by covering diverse domains, including \textbf{natural science}, \textbf{cultural common sense}, and \textbf{spatio-temporal-logic reasoning}. 
Besides, we also provide both Chinese and English versions for each case to evaluate the cross-lingual instruction-following abilities.

To comprehensively assess model capabilities, we finally curate 1,220 high-quality cases in WiseEdit and evaluate 21 leading models (17 open-source, 4 closed-source).
Beyond conventional metrics~\cite{liu2025step1x}, we further introduce two novel dimensions: \textbf{Knowledge Fidelity} to verify the application of correct knowledge, and \textbf{Creative Fusion} to quantify the degree of editing creativity.
Extensive experiments reveal the limitations of current SoTA models in performing cognition- and creativity-informed image editing, highlighting bottlenecks in their knowledge-based cognitive reasoning and creative composition capabilities. Overall, our contributions are three-fold:
\begin{itemize}
    \item We are the first to break down instruction-based image editing into three interconnected steps. And we propose WiseEdit, a novel benchmark designed for cognition- and creativity-informed image editing with rich task depth.
    \item WiseEdit covers a wide breadth of knowledge, including declarative, procedural, and metacognitive knowledge.
    \item With a comprehensive evaluation protocol, extensive experiments reveal the limitations of SoTA models when performing next-level intelligent image editing.

\end{itemize}

\begin{table*}[t!]
\centering 

\caption{\label{tab:compare}Comparison of open-source knowledge-informed reaonsing-based image editing benchmarks.}
\vspace{-1em}
\label{tab:image-editing}
\resizebox{1.0\textwidth}{!}{
\begin{tabular}{l|cc|cccc|ccc|cc}
\toprule
\multirow{2}{*}{\textbf{Benchmark}}  & \multirow{2}{*}{\textbf{\#Size}} & \multirow{2}{*}{\textbf{Multi-Img Input}} & \multicolumn{4}{c|}{\textbf{Task Depth}} & \multicolumn{3}{c|}{\textbf{Knowledge Breadth}} & \multicolumn{2}{c}{\textbf{Language}} \\ 
          &        &                 & Awareness & Interpretation & Imagination & Complex & Declarative & Procedural & Meta-cognitive &  ENG   &   CHN  \\ \hline

RiseBench~\cite{zhao2025envisioning} & 360 & 0\% & \ding{55}  & $\checkmark$ & \ding{55}& \ding{55}& $\checkmark$ & $\checkmark$& \ding{55} & $\checkmark$& \ding{55} \\
IntelligentBench~\cite{deng2025emerging} & 350 & 0\% & $\checkmark$  & $\checkmark$ & \ding{55}& \ding{55}& $\checkmark$ & $\checkmark$& \ding{55} & $\checkmark$& \ding{55} \\ 
KrisBench~\cite{wu2025kris}  & 1267 & 12\% & \ding{55}  & $\checkmark$ & \ding{55}& \ding{55}& $\checkmark$ & $\checkmark$& \ding{55} & $\checkmark$& \ding{55} \\ \hline
WiseEdit & 1220 & \textbf{26\%} & $\checkmark$ & $\checkmark$& $\checkmark$& $\checkmark$& $\checkmark$& $\checkmark$& $\checkmark$& $\checkmark$& $\checkmark$ \\
 \bottomrule
\end{tabular}%
}
\vspace{-1.2em}
\end{table*}

\section{Related Work}
\label{sec:formatting}

\paragraph{Image Editing Models.}

Early instruction-based image editing models~\cite{zhang2023magicbrush, zhao2024ultraedit, yu2025anyedit, flux-2-2025} focused on simple, localized modifications, such as object addition or removal.
Recent advancements have driven these models toward free-form unified image generation, primarily by  parsing the implicit intent within multimodal input via reasoning. 
Models like Bagel~\cite{deng2025emerging}, OmniGen2~\cite{wu2025omnigen2}, and Qwen-Image-Edit~\cite{wu2025qwen} achieve this through a unified visual comprehension-generation architecture that integrates a VLM with a diffusion model, enabling powerful reasoning-based editing. 
Furthermore, closed-source systems~\cite{seedream2025seedream, openai2024gpt4ocard, gemini2_5_flash_image, nanobananapro} further push performance boundaries by facilitating cognition- and creativity-informed image editing with strong knowledge-based reasoning and compositional creation.

\paragraph{Image Editing Benchmarks.} 

Existing image editing benchmarks~\cite{zhang2023magicbrush, ju2023direct, liu2025step1x} often feature overly simplistic intentions in the instructions and lack explicit modeling of the knowledge structures, thus failing to accurately assess advanced models' true capabilities.
Recently, inspired by the benchmarks~\cite{niu2025wise, li2025easier} for text-to-image generation tasks, new benchmarks such as IntelligentBench~\cite{deng2025emerging}, RiseBench~\cite{zhao2025envisioning}, and KrisBench~\cite{wu2025kris} have emerged to specifically gauge editing performance requiring knowledge-based reasoning.
However, these benchmarks remain limited as they typically use single-image inputs, primarily focus on the interpretation step while neglecting the step of awareness, imagination and other complex scenarios.
Moreover, they exclude the assessment of meta-cognitive knowledge, also restricting instructions to English language,
In contrast, WiseEdit offers a more comprehensive evaluation of cognition- and creativity-informed image editing from these perspectives, as detailed in the comparative analysis in Table~\ref{tab:compare}.

\section{WiseEdit}

\begin{figure}[t]
\includegraphics[width=\linewidth]{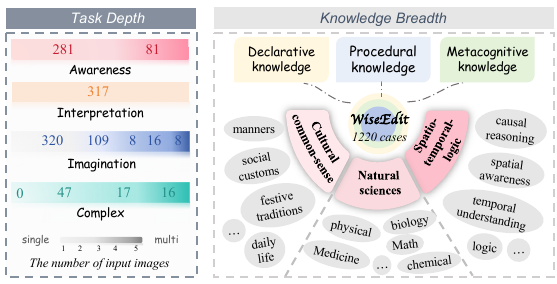}
\vspace{-1.5em}
\centering\caption{Taxonomy of WiseEdit (WiseEdit-Complex).}
\label{fig:tax}
\vspace{-1.7em}
\end{figure}

In this section, we introduce WiseEdit, a knowledge-intensive benchmark designed to evaluate cognition- and creativity-informed image editing with next-level intelligent. It features both high task depth and rich knowledge breadth, with diverse task formats and automated evaluation. The taxonomy of WiseEdit is presented in Fig.~\ref{fig:tax}.

\subsection{Benchmark Construction }

A critical leap in cognition- and creativity-informed intelligent image editing is the models' growing ability to infer implicit intent from multimodal input, and enable sophisticated creation that embodies this intent. To better measure this, we break down image editing into three interconnected steps, designing high-difficulty tasks for each, resulting in high task depth and diverse task formats within WiseEdit.

\begin{figure*}[t]
\includegraphics[width=\linewidth]{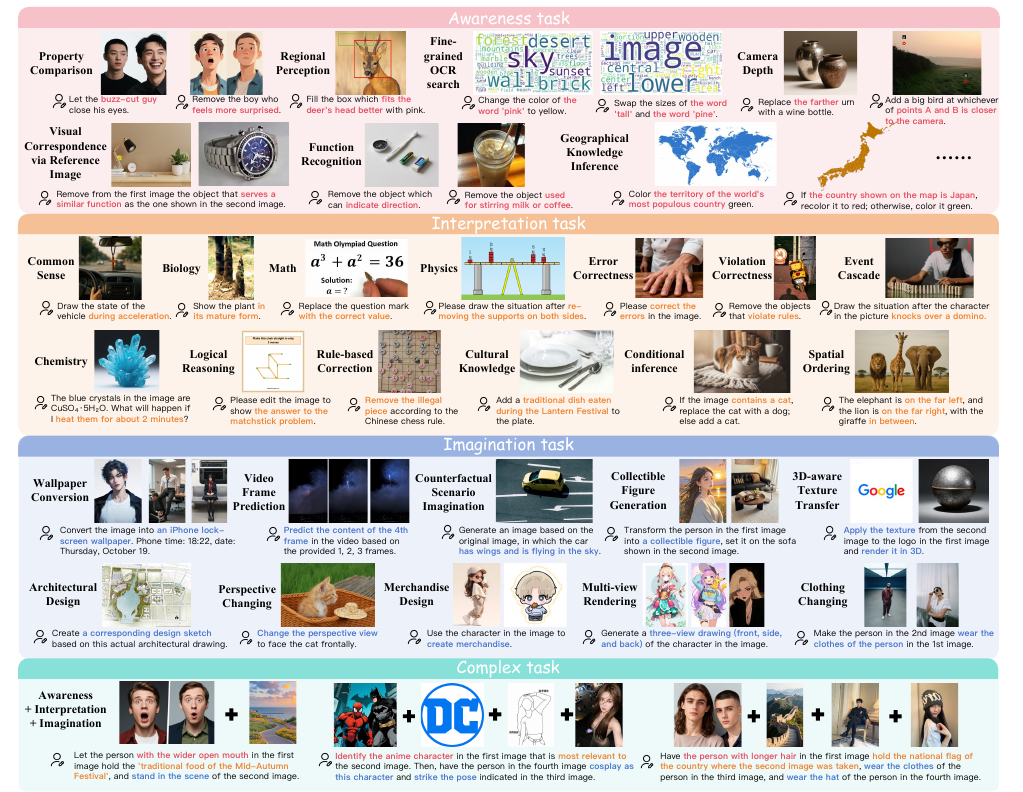}
\vspace{-1.7em}
\centering\caption{Examples of each task type in WiseEdit and WiseEdit-Complex.}
\label{fig:examples}
\vspace{-1.7em}
\end{figure*}

\paragraph{Task Depth. } 

In Section $\ref{sec:intro}$, we decompose the image editing process into three cascaded steps, \textit{i.e.}, awareness, interpretation, and imagination. Correspondingly, we first increase the task difficulty for each step and prevent the models from trivially completing the respective step:

\textbf{Awareness Task.} 
Awareness is the preliminary step in image editing where the AI model functions as a visual observer to establish selective attention on the input image. 
We design awareness tasks, challenging the model to deduce the target modification area without explicit spatial information provided in the instruction. 
This necessitates the application of reasoning capabilities, such as comparative reasoning (\textit{e.g.}, ``\textit{remove the happiest person}''), parsing the referent of an indirect reference (\textit{e.g.}, ``\textit{highlight the country of Shakespeare's homeland on ma}p''), or establishing visual correspondence from another reference image.

\textbf{Interpretation Task.}
Interpretation serves as the cognitive step, decoding the editing instructions into  executable perception-level visual changes.
In Interpretation task, instructions often do not explicitly state the required modification. The models must leverage  world knowledge to parsing the implicit intent into a directly executable action sequence.
For example, an implicit instruction might require \textit{rectifying an unknown error} in an anomalous image or \textit{performing temporal shift} of the image scene.
Even seemingly straightforward instructions necessitate models considering the \textit{cascaded reactions} that their execution might trigger.

\textbf{Imagination Task.}
Imagination serves as the generative step, rendering the visual edits parsed previously onto the target regions of the original image.
For this task, we introduce highly challenging subject-driven generation tasks that require models to perform imaginative and artistic creations while preserving the subject's identity. This involves sophisticated transformations like altering the subject's clothing, pose, or viewpoint, generating complex compositions of multiple objects, and even counter-intuitive creations (\textit{e.g., transforming human into construction scaffolding}).

Building upon this, we further designed WiseEdit-Complex, where none of the three editing steps can be easily finished, requiring the model to fully leverage its capacity for complex reasoning and creative generation. Fig.~\ref{fig:examples} presents examples of the each task type within WiseEdit(-Complex), with more details available in Appendix~\ref{app:1}.

\paragraph{Task Format.}

Existing image editing benchmarks typically use a single text-image input pair, which often limits meeting practical user needs because text alone may fail to convey abstract attributes (like texture), and some requests inherently require merging concepts from multiple images. 
To this end, WiseEdit expands the task format to include a large volume of cases with multiple input images, and instructions utilize ordinal terms to indicate the specific image.
Moreover, input images serve distinct roles: some are the targets for editing, while others function as references providing necessary abstract attributes or concrete objects to convey intent. Models must determine the specific role of each input image based on the instruction description.

Based on this, WiseEdit contains 1,220 high-quality test cases, with each instruction provided in both English and Chinese. See details on data collection in Appendix~\ref{app:2}.

\subsection{Knowledge-Intensive Evaluation in WiseEdit} 

Cognition- and creativity-informed intelligent image editing also requires the possession of robust world knowledge to facilitate superior image generation.
To measure this key ability, we incorporates an extensive set of world-knowledge-informed cases in WiseEdit, which cover a broad knowledge breadth with three knowledge types:

\textbf{Declarative knowledge}, often described as ``\textbf{\textit{knowing what}}'', encompasses facts and concepts that can be explicitly stated and defined. In the context of knowledge-intensive image editing, it not only includes observable perceptual cues but also integrates a higher-order understanding that connects perceptual information to generalizable principles and conceptual facts. For example, given an instruction ``\textit{add a toy animal that is a national treasure of China}'', the model must possess the declarative knowledge that the national treasure animal of China is the panda.

\textbf{Procedural knowledge}, defined as ``\textbf{\textit{knowing how}}'', encompasses the expertise and skills necessary to perform a task, which is often dynamic and difficult to explain verbally. 
For image editing, it requires a model to perform multi-step task decomposition to ascertain the correct procedure for executing a desired change. 
For example, when required to \textit{convert a watercolor painting into a line drawing}, the model relies on procedural knowledge to complete the process systematically.

\textbf{Metacognitive knowledge} is ``\textbf{\textit{knowing about knowing}}'',  requiring self-awareness and self-regulation during image editing. 
It dictates the high-level management of when to invoke declarative or procedural knowledge, and how to effectively combine them. 
For example, executing a conditional instruction ``\textit{have the girl hold the longest object in the second image; if the object can be used to brush teeth, let her face the camera, otherwise turn her back to the camera}'', requires that the model knows when and why to apply the necessary declarative or procedural steps.

Based on this, we also systematically encompass three critical knowledge domains—\textbf{Cultural Common Sense} (\textit{e.g.}, social customs, daily life), \textbf{Natural Sciences} (\textit{e.g.}, physical laws, chemical reactions), and \textbf{Spatio-Temporal-Logic} (\textit{e.g.}, causality, spatial arrangement, and temporal progression)—to assess culturally-appropriate, physically-consistent, and logically-coherent image editing.

\subsection{Evaluation Pipeline}
\label{sec:3.3}

To comprehensively evaluate the performance of SoTA models on WiseEdit, 
we employ GPT-4o as the automatic evaluator. 
Beyond widely used metrics such as Instruction Following, Detail Preserving and Visual Quality~\cite{liu2025step1x,gu2024multi}, we introduce two novel, essential metrics: Knowledge Fidelity and Creative Fusion. 
The former is designed to verify the application of correct knowledge, while the latter quantifies the degree of creativity exhibited by the models. 
Furthermore, for knowledge-informed cases, we provide a knowledge hint to precisely elucidate the explicit intent of the editing, assisting in the evaluation of Knowledge Fidelity.

\textbf{Instruction Following} ($\mathcal{IF}$) evaluates whether models accurately follow the editing request. \textbf{Detail Preserving} ($\mathcal{DP}$) evaluates whether models faithfully preserve the parts not related to the instruction.
\textbf{Visual Quality} ($\mathcal{VP}$) evaluates the perceptual quality of the generated image, such as its naturalness and any presence of artifacts.

\textbf{Knowledge Fidelity} ($\mathcal{KF}$) evaluates whether the edited image accurately reflects real-world logic and knowledge, including physical, biological, chemical, cultural, commonsense correctness, and so on, based on the provided knowledge hint which serves as a guiding reference.

\textbf{Creative Fusion} ($\mathcal{CF}$) evaluates conceptual novelty, expressive transformation, and imaginative depth of the edited image relative to the originals, gauging its demonstration of human-like creative thinking for imagination tasks.

All test cases are evaluated using $\mathcal{IF}$, $\mathcal{DP}$ and $\mathcal{VQ}$. 
with $\mathcal{KF}$ only for  Awareness, Interpretation, and Wise-Complex tasks,  $\mathcal{CF}$ for Imagination and Wise-Complex tasks. 
Each metrics is rated on a 1–10 scale based on carefully custom-crafted prompts to ensure precise scoring. The overall performance metric is calculated as 
$\mathbf{AVG} = (\mathcal{IF}+\mathcal{DP}+\mathcal{VQ}+\alpha\cdot \mathcal{KF} +\beta\cdot \mathcal{CF})/(3+\alpha+\beta)$,
where $\alpha$ and $\beta$ are set to $1$ if the respective metric ($\mathcal{KF}$ or $\mathcal{CF}$) is required for the task, and $0$ otherwise. See more details in Appendix~\ref{app:3}.

\section{Experiments}

\begin{table*}[t]
    \centering
    \caption{ \label{tab:main} Main results on WiseEdit with both English and Chinese task version. The best results are marked in \textbf{bold} for {\color[HTML]{F88825} open-} and {\color[HTML]{319B62} closed-}models, respectively.}
    \vspace{-1em}
    \resizebox{1.0\linewidth}{!}{
        \begin{tabular}{lc|ccccc|ccccc|ccccc|c}
            \toprule
             \multirow{2}{*}{Model}    & \multirow{2}{*}{Multi-Img}&\multicolumn{5}{c}{\textbf{Awareness Task}}  &
                    \multicolumn{5}{c}{\textbf{Interpretation Task}}  &
                    \multicolumn{5}{c}{\textbf{Imagination Task}}  &
                    
                   \textbf{Overall} \\
            &  & $\mathcal{IF}\uparrow$ & $\mathcal{DP}\uparrow$ & $\mathcal{VQ}\uparrow$ & $\mathcal{KF}\uparrow$ & $\cellcolor[HTML]{E2E2E2}\mathbf{AVG}$ & $\mathcal{IF}\uparrow$ & $\mathcal{DP}\uparrow$ & $\mathcal{VQ}\uparrow$ & $\mathcal{KF}\uparrow$ & $\cellcolor[HTML]{E2E2E2}\mathbf{AVG}$& $\mathcal{IF}\uparrow$ & $\mathcal{DP}\uparrow$ & $\mathcal{VQ}\uparrow$ & $\mathcal{CF}\uparrow$ & $\cellcolor[HTML]{E2E2E2}\mathbf{AVG}$& $\cellcolor[HTML]{C8C8C8}\mathbf{AVG}$\\ \hline
            \multicolumn{18}{c}{\textit{English Version}} \\ \hline
InstructPix2Pix~\cite{brooks2023instructpix2pix} & \ding{55} & 24.6 & 33.7 & 50.6 & 26.4 & \cellcolor[HTML]{E2E2E2}33.8 & 20.7 & 50.3 & 66.9 & 23.6 & \cellcolor[HTML]{E2E2E2}40.4 & 17.0 & 29.9 & 41.2 & 27.4 & \cellcolor[HTML]{E2E2E2}28.8 & \cellcolor[HTML]{C8C8C8}34.4 \\
MagicBrush~\cite{zhang2023magicbrush} & \ding{55}& 27.2 & 43.4 & 53.3 & 27.1 & \cellcolor[HTML]{E2E2E2}37.8 & 16.8 & 50.3 & 63.2 & 22.2 & \cellcolor[HTML]{E2E2E2}38.1 & 18.0 & 36.9 & 44.8 & 22.3 & \cellcolor[HTML]{E2E2E2}30.5 & \cellcolor[HTML]{C8C8C8}35.5 \\
OmniGen~\cite{xiao2025omnigen} & $\checkmark$ & 35.0 & 42.0 & 46.7 & 37.4 & \cellcolor[HTML]{E2E2E2}40.3 & 19.0 & 34.8 & 40.3 & 21.5 & \cellcolor[HTML]{E2E2E2}28.9 & 42.2 & 35.1 & 46.0 & 38.7 & \cellcolor[HTML]{E2E2E2}40.5 & \cellcolor[HTML]{C8C8C8}36.6 \\
AnyEdit~\cite{yu2025anyedit} & $\checkmark$ & 25.0 & 54.6 & 61.3 & 26.3 & \cellcolor[HTML]{E2E2E2}41.8 & 15.9 & 61.2 & 62.0 & 20.2 & \cellcolor[HTML]{E2E2E2}39.8 & 9.1 & 49.7 & 50.9 & 16.5 & \cellcolor[HTML]{E2E2E2}31.5 & \cellcolor[HTML]{C8C8C8}37.7 \\
UltraEdit~\cite{zhao2024ultraedit} & \ding{55}& 26.5 & 42.5 & 53.1 & 33.9 & \cellcolor[HTML]{E2E2E2}39.0 & 24.3 & 61.7 & 73.6 & 26.7 & \cellcolor[HTML]{E2E2E2}46.6 & 20.7 & 31.7 & 45.8 & 27.5 & \cellcolor[HTML]{E2E2E2}31.5 & \cellcolor[HTML]{C8C8C8}39.0 \\
ICEdit~\cite{zhang2025context} & \ding{55}& 26.1 & 42.2 & 61.2 & 31.8 & \cellcolor[HTML]{E2E2E2}40.4 & 21.4 & 48.3 & 81.5 & 24.9 & \cellcolor[HTML]{E2E2E2}44.0 & 21.5 & 40.6 & 54.0 & 25.0 & \cellcolor[HTML]{E2E2E2}35.3 & \cellcolor[HTML]{C8C8C8}39.9 \\
FLUX.1 Kontext Dev~\cite{labs2025flux} & \ding{55}& 31.4 & 52.0 & 55.0 & 35.5 & \cellcolor[HTML]{E2E2E2}43.5 & 27.5 & 62.2 & 69.6 & 29.0 & \cellcolor[HTML]{E2E2E2}47.1 & 39.1 & 47.1 & 43.4 & 27.1 & \cellcolor[HTML]{E2E2E2}39.2 & \cellcolor[HTML]{C8C8C8}43.2   \\ 

FLUX.2 Dev~\cite{flux-2-2025} & $\checkmark$ & 42.6 & 63.3 & 78.4 & 53.3 & \cellcolor[HTML]{E2E2E2}59.4 & 35.4 & 75.0 & 85.6 & 37.6 & \cellcolor[HTML]{E2E2E2}58.4 & \color[HTML]{F88825}\textbf{73.6} & \color[HTML]{F88825}\textbf{70.7} & \color[HTML]{F88825}\textbf{82.1} & \color[HTML]{F88825}\textbf{43.6} & \cellcolor[HTML]{E2E2E2}\color[HTML]{F88825}\textbf{67.5} & \cellcolor[HTML]{C8C8C8}\color[HTML]{F88825}\textbf{61.8} \\
\hline
Janus-4o~\cite{chen2025sharegpt} &\ding{55} & 34.7 & 37.0 & 45.9 & 36.2 & \cellcolor[HTML]{E2E2E2}38.5 & 27.2 & 43.8 & 53.6 & 28.2 & \cellcolor[HTML]{E2E2E2}38.2 & 28.2 & 37.6 & 42.0 & 25.5 & \cellcolor[HTML]{E2E2E2}33.3 & \cellcolor[HTML]{C8C8C8}36.7 \\
UniWorld-V1~\cite{lin2025uniworld} & $\checkmark$ & 31.5 & 48.9 & 58.8 & 38.6 & \cellcolor[HTML]{E2E2E2}44.5 & 18.1 & 44.5 & 58.1 & 22.5 & \cellcolor[HTML]{E2E2E2}35.8 & 30.3 & 50.3 & 64.2 & 27.5 & \cellcolor[HTML]{E2E2E2}43.1 & \cellcolor[HTML]{C8C8C8}41.1 \\
HiDream-E1~\cite{cai2025hidream} & \ding{55}& 29.7 & 41.2 & 56.3 & 32.0 & \cellcolor[HTML]{E2E2E2}39.8 & 26.7 & 53.6 & 68.4 & 29.6 & \cellcolor[HTML]{E2E2E2}44.6 & 39.6 & 40.1 & 49.9 & 29.6 & \cellcolor[HTML]{E2E2E2}39.8 & \cellcolor[HTML]{C8C8C8}41.4 \\
OmniGen2~\cite{wu2025omnigen2} & $\checkmark$ & 35.0 & 64.0 & 75.4 & 41.3 & \cellcolor[HTML]{E2E2E2}53.9 & 18.9 & 56.9 & 64.9 & 23.5 & \cellcolor[HTML]{E2E2E2}41.1 & 42.0 & 64.4 & 74.6 & 31.8 & \cellcolor[HTML]{E2E2E2}53.2 & \cellcolor[HTML]{C8C8C8}49.4 \\
Step1X-Edit-v1p2~\cite{liu2025step1x} &  \ding{55} & 39.8 & 53.5 & 61.3 & 44.4 & \cellcolor[HTML]{E2E2E2}49.7 & 35.7 & 73.0 & 75.2 & 38.2 & \cellcolor[HTML]{E2E2E2}55.5 & 44.7 & 49.4 & 50.3 & 28.4 & \cellcolor[HTML]{E2E2E2}43.2 & \cellcolor[HTML]{C8C8C8}49.5 \\
Echo-4o~\cite{ye2025echo} & $\checkmark$& 47.6 & 63.0 & 75.4 & 51.7 & \cellcolor[HTML]{E2E2E2}59.4 & 30.8 & 71.0 & 80.4 & 32.9 & \cellcolor[HTML]{E2E2E2}53.8 & 63.4 & 62.4 & 73.7 & 41.2 & \cellcolor[HTML]{E2E2E2}60.2 & \cellcolor[HTML]{C8C8C8}57.8 \\
Bagel~\cite{deng2025emerging} & $\checkmark$ & 46.2 & 71.0 & 75.8 & 50.8 & \cellcolor[HTML]{E2E2E2}61.0 & \color[HTML]{F88825} \textbf{38.6} & 72.1 & 78.8 & \color[HTML]{F88825} \textbf{39.5} & \cellcolor[HTML]{E2E2E2}57.3 & 62.8 & 68.5 & 74.5 & 40.7 & \cellcolor[HTML]{E2E2E2}61.6 & \cellcolor[HTML]{C8C8C8}60.0 \\
Uni-CoT~\cite{qin2025uni} & $\checkmark$ & 46.0 & 69.1 & 77.8 & 51.6 & \cellcolor[HTML]{E2E2E2}61.1 & 36.9 & 70.1 & 76.3 & 38.6 & \cellcolor[HTML]{E2E2E2}55.5 & 67.6 & 64.3 & 79.6 & 42.9 & \cellcolor[HTML]{E2E2E2}63.6 & \cellcolor[HTML]{C8C8C8}60.1 \\
Qwen-Image-Edit~\cite{wu2025qwen} & $\checkmark$ & \color[HTML]{F88825} \textbf{48.1} & 69.0 & 79.5 & \color[HTML]{F88825} \textbf{53.6} & \cellcolor[HTML]{E2E2E2}62.5 & 32.1 & 69.7 & 80.6 & 34.2 & \cellcolor[HTML]{E2E2E2}54.1 & 67.1 & 66.8 & 79.2 & 42.3 & \cellcolor[HTML]{E2E2E2}63.8 & \cellcolor[HTML]{C8C8C8}60.2 \\
DreamOmni2~\cite{xia2025dreamomni2} & $\checkmark$ & 43.3 & \color[HTML]{F88825} \textbf{74.4} & \color[HTML]{F88825} \textbf{85.0} & 51.2 & \cellcolor[HTML]{E2E2E2}\color[HTML]{F88825} \textbf{63.5} & 34.3 & \color[HTML]{F88825} \textbf{81.7} & \color[HTML]{F88825} \textbf{88.1} & 35.9 & \cellcolor[HTML]{E2E2E2}\color[HTML]{F88825} \textbf{60.0} & 50.6 & 64.9 & 81.9 & 35.3 & \cellcolor[HTML]{E2E2E2}58.2 & \cellcolor[HTML]{C8C8C8}60.6 \\ \hline
Nano Banana~\cite{gemini2_5_flash_image} &  $\checkmark$& 70.6 & 85.7 & 86.8 & 75.2 & \cellcolor[HTML]{E2E2E2}79.6 & 63.4 & 84.9 & 91.4 & 61.5 & \cellcolor[HTML]{E2E2E2}75.3 & 75.3 & 73.8 & 87.3 & 44.3 & \cellcolor[HTML]{E2E2E2}70.2 & \cellcolor[HTML]{C8C8C8}75.0 \\
Seedream 4.0~\cite{seedream2025seedream} &  $\checkmark$ & 70.8 & 78.1 & 86.6 & 74.6 & \cellcolor[HTML]{E2E2E2}77.5 & 63.7 & 80.1 & 90.6 & 64.2 & \cellcolor[HTML]{E2E2E2}74.6 & 82.2 & 77.8 & 86.9 & 47.0 & \cellcolor[HTML]{E2E2E2}73.5 & \cellcolor[HTML]{C8C8C8}75.2 \\
GPT-image-1~\cite{openai2024gpt4ocard} & $\checkmark$ & 78.5 & 85.8 & \color[HTML]{319B62} \textbf{88.0} & 81.2 & \cellcolor[HTML]{E2E2E2}83.3 & 62.9 & 82.9 & \color[HTML]{319B62} \textbf{93.0} & 60.8 & \cellcolor[HTML]{E2E2E2}74.9 & 84.4 & 76.2 & \color[HTML]{319B62} \textbf{89.2} & 48.4 & \cellcolor[HTML]{E2E2E2}74.6 & \cellcolor[HTML]{C8C8C8}77.6 \\ 
Nano Banana Pro~\cite{nanobananapro} &  $\checkmark$ & \color[HTML]{319B62} \textbf{85.4} & \color[HTML]{319B62} \textbf{88.6} & 83.9 & \color[HTML]{319B62} \textbf{91.4} & \cellcolor[HTML]{E2E2E2}\color[HTML]{319B62} \textbf{87.3} & \color[HTML]{319B62} \textbf{76.0} & \color[HTML]{319B62} \textbf{89.1} & 92.3 & \color[HTML]{319B62} \textbf{75.8} & \cellcolor[HTML]{E2E2E2}\color[HTML]{319B62} \textbf{83.3} & \color[HTML]{319B62} \textbf{86.6} & \color[HTML]{319B62} \textbf{79.5} & 88.8 & \color[HTML]{319B62} \textbf{51.5} & \cellcolor[HTML]{E2E2E2}\color[HTML]{319B62} \textbf{76.6} & \cellcolor[HTML]{C8C8C8}\color[HTML]{319B62} \textbf{82.4} \\ \hline
    \multicolumn{18}{c}{\textit{Chinese Version}} \\  \hline

InstructPix2Pix~\cite{brooks2023instructpix2pix} & \ding{55}& 14.2 & 51.0 & 58.6 & 18.1 & \cellcolor[HTML]{E2E2E2}35.5 & 13.0 & 55.7 & 65.2 & 16.0 & \cellcolor[HTML]{E2E2E2}37.5 & 3.7 & 52.0 & 53.2 & 9.0 & \cellcolor[HTML]{E2E2E2}29.5 & \cellcolor[HTML]{C8C8C8}34.1 \\
MagicBrush~\cite{zhang2023magicbrush} & \ding{55}& 15.0 & 43.8 & 52.9 & 17.8 & \cellcolor[HTML]{E2E2E2}32.4 & 10.1 & 40.5 & 59.4 & 13.5 & \cellcolor[HTML]{E2E2E2}30.9 & 5.4 & 39.6 & 49.2 & 12.7 & \cellcolor[HTML]{E2E2E2}26.8 & \cellcolor[HTML]{C8C8C8}30.0 \\
OmniGen~\cite{xiao2025omnigen} & $\checkmark$& 16.3 & 42.6 & 60.1 & 22.6 & \cellcolor[HTML]{E2E2E2}35.4 & 13.4 & 30.9 & 49.2 & 14.8 & \cellcolor[HTML]{E2E2E2}27.1 & 15.4 & 27.6 & 51.3 & 32.3 & \cellcolor[HTML]{E2E2E2}31.7 & \cellcolor[HTML]{C8C8C8}31.3 \\
AnyEdit~\cite{yu2025anyedit} & $\checkmark$& 17.5 & 55.3 & 55.7 & 19.9 & \cellcolor[HTML]{E2E2E2}37.1 & 11.4 & 51.8 & 58.0 & 16.0 & \cellcolor[HTML]{E2E2E2}34.3 & 7.3 & 47.4 & 52.3 & 15.4 & \cellcolor[HTML]{E2E2E2}30.6 & \cellcolor[HTML]{C8C8C8}34.0 \\
UltraEdit~\cite{zhao2024ultraedit} & \ding{55}& 16.9 & 58.5 & 62.1 & 21.2 & \cellcolor[HTML]{E2E2E2}39.7 & 17.4 & 74.2 & 78.9 & 19.2 & \cellcolor[HTML]{E2E2E2}47.4 & 9.3 & 42.3 & 51.8 & 14.8 & \cellcolor[HTML]{E2E2E2}29.5 & \cellcolor[HTML]{C8C8C8}38.9 \\
ICEdit~\cite{zhang2025context} & \ding{55}& 12.9 & 29.1 & 63.6 & 17.4 & \cellcolor[HTML]{E2E2E2}30.8 & 11.5 & 43.5 & 81.1 & 17.0 & \cellcolor[HTML]{E2E2E2}38.3 & 5.1 & 37.4 & 56.5 & 16.7 & \cellcolor[HTML]{E2E2E2}28.9 & \cellcolor[HTML]{C8C8C8}32.7 \\
FLUX.1 Kontext Dev~\cite{labs2025flux} & \ding{55}& 16.5 & 48.4 & 52.1 & 19.1 & \cellcolor[HTML]{E2E2E2}34.0 & 16.8 & 58.4 & 66.1 & 22.2 & \cellcolor[HTML]{E2E2E2}40.9 & 9.2 & 41.9 & 43.3 & 10.6 & \cellcolor[HTML]{E2E2E2}26.3 & \cellcolor[HTML]{C8C8C8}33.7 \\ 

FLUX.2 Dev~\cite{flux-2-2025} &$\checkmark$ & 43.0 & 60.6 & 79.5 & 51.4 & \cellcolor[HTML]{E2E2E2}58.6 & 34.3 & 73.7 & 83.1 & 36.0 & \cellcolor[HTML]{E2E2E2}\color[HTML]{F88825} \textbf{56.8} & \color[HTML]{F88825} \textbf{75.5} & \color[HTML]{F88825} \textbf{74.4} & 82.8 & \color[HTML]{F88825} \textbf{42.6} & \cellcolor[HTML]{E2E2E2}\color[HTML]{F88825} \textbf{68.8} & \cellcolor[HTML]{C8C8C8}\color[HTML]{F88825} \textbf{61.4} \\
\hline
Janus-4o~\cite{chen2025sharegpt} & \ding{55}& 31.1 & 38.6 & 46.3 & 34.1 & \cellcolor[HTML]{E2E2E2}37.5 & 23.9 & 45.9 & 55.7 & 25.8 & \cellcolor[HTML]{E2E2E2}37.8 & 25.4 & 36.9 & 41.8 & 22.1 & \cellcolor[HTML]{E2E2E2}31.5 & \cellcolor[HTML]{C8C8C8}35.6 \\
UniWorld-V1~\cite{lin2025uniworld} & $\checkmark$& 18.4 & 49.0 & 60.0 & 26.2 & \cellcolor[HTML]{E2E2E2}38.4 & 13.3 & 48.8 & 59.7 & 16.9 & \cellcolor[HTML]{E2E2E2}34.7 & 17.9 & 54.8 & 68.9 & 18.0 & \cellcolor[HTML]{E2E2E2}39.9 & \cellcolor[HTML]{C8C8C8}37.7 \\
HiDream-E1~\cite{cai2025hidream} & \ding{55}& 28.2 & 37.6 & 51.4 & 32.4 & \cellcolor[HTML]{E2E2E2}37.4 & 25.4 & 47.1 & 63.6 & 29.7 & \cellcolor[HTML]{E2E2E2}41.5 & 32.5 & 39.2 & 47.5 & 27.4 & \cellcolor[HTML]{E2E2E2}36.6 & \cellcolor[HTML]{C8C8C8}38.5 \\
OmniGen2~\cite{wu2025omnigen2} &  $\checkmark$& 35.1 & 57.9 & 72.4 & 41.0 & \cellcolor[HTML]{E2E2E2}51.6 & 19.1 & 57.1 & 64.8 & 23.0 & \cellcolor[HTML]{E2E2E2}41.0 & 45.5 & 64.0 & 72.0 & 33.8 & \cellcolor[HTML]{E2E2E2}53.8 & \cellcolor[HTML]{C8C8C8}48.8 \\
Step1X-Edit-v1p2~\cite{liu2025step1x} & \ding{55}& 38.6 & 55.6 & 59.5 & 42.0 & \cellcolor[HTML]{E2E2E2}48.9 & 37.0 & 77.5 & 76.8 & 35.9 & \cellcolor[HTML]{E2E2E2}\color[HTML]{F88825} \textbf{56.8} & 45.7 & 48.3 & 51.3 & 27.0 & \cellcolor[HTML]{E2E2E2}43.1 & \cellcolor[HTML]{C8C8C8}49.6 \\
Echo-4o~\cite{ye2025echo} & $\checkmark$ & 47.9 & 59.9 & 73.1 & 55.0 & \cellcolor[HTML]{E2E2E2}59.0 & 31.9 & 74.5 & 77.4 & 32.6 & \cellcolor[HTML]{E2E2E2}54.1 & 62.8 & 64.2 & 75.1 & 41.5 & \cellcolor[HTML]{E2E2E2}60.9 & \cellcolor[HTML]{C8C8C8}58.0 \\
Bagel~\cite{deng2025emerging} & $\checkmark$ & \color[HTML]{F88825} \textbf{48.5} & 71.3 & 76.8 & 52.1 & \cellcolor[HTML]{E2E2E2}62.2 & 36.5 & 68.7 & 75.0 & \color[HTML]{F88825} \textbf{38.5} & \cellcolor[HTML]{E2E2E2}54.7 & 63.5 & 68.3 & 75.3 & 39.7 & \cellcolor[HTML]{E2E2E2}61.7 & \cellcolor[HTML]{C8C8C8}59.5 \\
Uni-CoT~\cite{qin2025uni} & $\checkmark$ & 46.2 & 70.0 & 80.7 & \color[HTML]{F88825} \textbf{53.6} & \cellcolor[HTML]{E2E2E2}\color[HTML]{F88825} \textbf{62.6} & \color[HTML]{F88825} \textbf{37.4} & 71.5 & 79.2 & 36.6 & \cellcolor[HTML]{E2E2E2}56.2 & 65.5 & 65.1 & 79.7 & 41.6 & \cellcolor[HTML]{E2E2E2}63.0 & \cellcolor[HTML]{C8C8C8} 60.6 \\
Qwen-Image-Edit~\cite{wu2025qwen} & $\checkmark$ & 45.0 & 67.3 & 79.9 & 52.9 & \cellcolor[HTML]{E2E2E2}61.3 & 35.8 & 74.0 & 80.7 & 36.1 & \cellcolor[HTML]{E2E2E2}56.6 & 66.3 & 67.2 & 80.0 & 41.7 & \cellcolor[HTML]{E2E2E2}63.8 &  \cellcolor[HTML]{C8C8C8}60.6 \\
DreamOmni2~\cite{xia2025dreamomni2} & $\checkmark$ & 31.9 & \color[HTML]{F88825} \textbf{78.7} & \color[HTML]{F88825} \textbf{85.4} & 38.4 & \cellcolor[HTML]{E2E2E2}58.6 & 24.0 & \color[HTML]{F88825} \textbf{80.1} & \color[HTML]{F88825} \textbf{86.2} & 27.5 & \cellcolor[HTML]{E2E2E2}54.4 & 38.5 & 69.4 & \color[HTML]{F88825} \textbf{84.8} & 27.2 & \cellcolor[HTML]{E2E2E2}55.0 & \cellcolor[HTML]{C8C8C8}56.0 \\ \hline
Nano Banana~\cite{gemini2_5_flash_image} & $\checkmark$ & 71.8 & 83.8 & 86.5 & 70.7 & \cellcolor[HTML]{E2E2E2}78.2 & 67.9 & \color[HTML]{319B62} \textbf{84.5} & 91.4 & 63.7 & \cellcolor[HTML]{E2E2E2}76.9 & 76.0 & 75.8 & 87.3 & 43.7 & \cellcolor[HTML]{E2E2E2}70.7 & \cellcolor[HTML]{C8C8C8}75.3 \\
Seedream 4.0~\cite{seedream2025seedream} & $\checkmark$ & 69.1 & 79.0 & 84.4 & 72.0 & \cellcolor[HTML]{E2E2E2}76.1 & 62.2 & 80.3 & 89.9 & 59.9 & \cellcolor[HTML]{E2E2E2}73.1 & 79.8 & \color[HTML]{319B62} \textbf{79.7} & 86.5 & 46.4 & \cellcolor[HTML]{E2E2E2}73.1 & \cellcolor[HTML]{C8C8C8}74.1 \\
GPT-image-1~\cite{openai2024gpt4ocard} & $\checkmark$& 77.0 & 80.7 & \color[HTML]{319B62} \textbf{86.6} & 80.7 & \cellcolor[HTML]{E2E2E2}81.2 & 61.4 & 82.6 & \color[HTML]{319B62} \textbf{93.8} & 61.2 & \cellcolor[HTML]{E2E2E2}74.8 & 78.8 & 73.3 & \color[HTML]{319B62} \textbf{89.6} & 48.3 & \cellcolor[HTML]{E2E2E2}72.5 & \cellcolor[HTML]{C8C8C8}76.2 \\
Nano Banana Pro~\cite{nanobananapro} & $\checkmark$ & \color[HTML]{319B62} \textbf{84.6} & \color[HTML]{319B62} \textbf{91.8} & 83.1 & \color[HTML]{319B62} \textbf{87.9} & \cellcolor[HTML]{E2E2E2}\color[HTML]{319B62} \textbf{86.9} & \color[HTML]{319B62} \textbf{74.2} & 83.9 & 91.3 & \color[HTML]{319B62} \textbf{74.6} & \cellcolor[HTML]{E2E2E2}\color[HTML]{319B62} \textbf{81.0} & \color[HTML]{319B62} \textbf{85.5} & 77.6 & 88.4 & \color[HTML]{319B62} \textbf{51.1} & \cellcolor[HTML]{E2E2E2}\color[HTML]{319B62} \textbf{75.6} & \cellcolor[HTML]{C8C8C8}\color[HTML]{319B62} \textbf{81.2} \\

            \bottomrule
        \end{tabular}}
    \vspace{-1.0em}
\end{table*}

\begin{table*}[t]
    \centering
    \caption{ \label{tab:complex} Main results on WiseEdit-Complex. We exclude models unable to handle multi-image inputs. }
    \vspace{-1em}
    \resizebox{0.8\linewidth}{!}{
        \begin{tabular}{l|cccccc|cccccc|c}
            \toprule
            \multirow{2}{*}{Model}       &\multicolumn{6}{c}{\textbf{\textit{English Version}}}  &
                    \multicolumn{6}{c}{\textbf{\textit{Chinese Version}}}  &
                  \textbf{Overall }
                    
                    \\
              & $\mathcal{IF}\uparrow$ & $\mathcal{DP}\uparrow$ & $\mathcal{VQ}\uparrow$ & $\mathcal{KF}\uparrow$ & $\mathcal{CF}\uparrow$ & $\cellcolor[HTML]{E2E2E2}\mathbf{AVG}$ & $\mathcal{IF}\uparrow$ & $\mathcal{DP}\uparrow$ & $\mathcal{VQ}\uparrow$ & $\mathcal{KF}\uparrow$  & $\mathcal{CF}\uparrow$ & $\cellcolor[HTML]{E2E2E2}\mathbf{AVG}$ & $\cellcolor[HTML]{C8C8C8}\mathbf{AVG}$\\ \hline
AnyEdit &  2.5 & 5.6 & 20.6 & 3.3 & 11.7 & \cellcolor[HTML]{E2E2E2}8.7 & 1.3 & 5.1 & 22.2 & 2.9 & 9.3 & \cellcolor[HTML]{E2E2E2}8.2 & \cellcolor[HTML]{C8C8C8}8.5 \\ 
OmniGen &  23.5 & 25.7 & 41.2 & 31.5 & 48.9 & \cellcolor[HTML]{E2E2E2}34.2 & 4.4 & 15.4 & 50.3 & 15.1 & 32.2 & \cellcolor[HTML]{E2E2E2}23.5 & \cellcolor[HTML]{C8C8C8}28.9 \\ 
FLUX.2 Dev &  42.3 & \color[HTML]{F88825} \textbf{68.5} & 75.6 & 56.5 & 49.1 & \cellcolor[HTML]{E2E2E2}\color[HTML]{F88825} \textbf{58.4} & \color[HTML]{F88825} \textbf{46.3} & \color[HTML]{F88825} \textbf{73.4} & \color[HTML]{F88825} \textbf{80.3} & \color[HTML]{F88825} \textbf{59.9} & \color[HTML]{F88825} \textbf{52.8} & \cellcolor[HTML]{E2E2E2}\color[HTML]{F88825} \textbf{62.6} & \cellcolor[HTML]{C8C8C8}\color[HTML]{F88825} \textbf{60.5} \\
\hline
UniWorld-V1 &  18.1 & 32.1 & 55.3 & 22.8 & 28.6 & \cellcolor[HTML]{E2E2E2}31.4 & 8.8 & 23.8 & 64.6 & 12.4 & 15.4 & \cellcolor[HTML]{E2E2E2}25.0 & \cellcolor[HTML]{C8C8C8}28.2 \\
OmniGen2 &  34.1 & 50.2 & 72.4 & 49.0 & 44.8 & \cellcolor[HTML]{E2E2E2}50.1 & 30.2 & 51.7 & 75.1 & 47.3 & 45.8 & \cellcolor[HTML]{E2E2E2}50.0 & \cellcolor[HTML]{C8C8C8}50.1 \\
DreamOmni2 &  34.8 & 62.3 & 78.8 & 46.0 & 41.0 & \cellcolor[HTML]{E2E2E2}52.6 & 36.6 & 51.6 & 79.1 & 49.6 & 35.5 & \cellcolor[HTML]{E2E2E2}50.4 & \cellcolor[HTML]{C8C8C8}51.5 \\
Echo-4o &  42.7 & 46.5 & 64.1 & 50.9 & 48.7 & \cellcolor[HTML]{E2E2E2}50.6 & 41.0 & 55.5 & 68.4 & 53.0 & 50.1 & \cellcolor[HTML]{E2E2E2}53.6 & \cellcolor[HTML]{C8C8C8}52.1 \\
Qwen-Image-Edit &  38.7 & 58.6 & \color[HTML]{F88825}\textbf{75.8} & 48.5 & 47.1 & \cellcolor[HTML]{E2E2E2}53.8 & 35.3 & 55.0 & 78.1 & 49.6 & 48.9 & \cellcolor[HTML]{E2E2E2}53.4 & \cellcolor[HTML]{C8C8C8}53.6 \\
Bagel &  \color[HTML]{F88825}\textbf{43.8} & 62.0 & 70.0 & 53.5 & 43.4 & \cellcolor[HTML]{E2E2E2}54.5 & 39.5 & 58.2 & 73.5 & 55.5 & 46.8 & \cellcolor[HTML]{E2E2E2}54.7 & \cellcolor[HTML]{C8C8C8}54.6 \\
Uni-CoT &  35.5 & 60.0 & 69.3 & \color[HTML]{F88825}\textbf{57.0} & \color[HTML]{F88825}\textbf{49.5} & \cellcolor[HTML]{E2E2E2}54.3 & 36.4 & 55.2 & 77.4 & 58.1 & 48.9 & \cellcolor[HTML]{E2E2E2}55.2 & \cellcolor[HTML]{C8C8C8}54.8 \\

\hline
Nano Banana & 53.8 & 75.2 & 82.7 & 82.4 & 53.7 & \cellcolor[HTML]{E2E2E2}69.6 & 53.3 & 71.3 & 79.9 & 77.4 & 51.1 & \cellcolor[HTML]{E2E2E2}66.6 & \cellcolor[HTML]{C8C8C8}68.1 \\
GPT-image-1 &  58.7 & 75.9 & \color[HTML]{319B62}\textbf{87.6} & 77.9 & 54.8 & \cellcolor[HTML]{E2E2E2}71.0 & 59.5 & 76.6 & \color[HTML]{319B62}\textbf{88.2} & 78.8 & 54.1 & \cellcolor[HTML]{E2E2E2}71.4 & \cellcolor[HTML]{C8C8C8}71.2 \\
Seedream 4.0 &  67.2 & 77.3 & 79.6 & \color[HTML]{319B62}\textbf{89.7} & 53.1 & \cellcolor[HTML]{E2E2E2}73.4 & 59.3 & 62.8 & 81.0 & 87.8 & 55.5 & \cellcolor[HTML]{E2E2E2}69.3 & \cellcolor[HTML]{C8C8C8}71.4 \\
Nano Banana Pro &  \color[HTML]{319B62} \textbf{68.1} & \color[HTML]{319B62} \textbf{78.1} & 86.7 & 88.1 & \color[HTML]{319B62} \textbf{56.6} & \cellcolor[HTML]{E2E2E2}\color[HTML]{319B62} \textbf{75.5} & \color[HTML]{319B62} \textbf{77.7} & \color[HTML]{319B62} \textbf{83.8} & 84.3 & \color[HTML]{319B62} \textbf{90.7} & \color[HTML]{319B62} \textbf{57.0} & \cellcolor[HTML]{E2E2E2}\color[HTML]{319B62} \textbf{78.7} & \cellcolor[HTML]{C8C8C8}\color[HTML]{319B62} \textbf{77.1} \\

            \bottomrule
        \end{tabular}}
    \vspace{-1.0em}
\end{table*}

\subsection{Experimental Setup.}

We evaluate 22 mainstream image editing models across architectures, covering both open- and closed-source models: 
\textbf{(1) Diffusion models:} InstructPix2Pix~\cite{brooks2023instructpix2pix}, MagicBrush~\cite{zhang2023magicbrush}, OmniGen~\cite{xiao2025omnigen}, AnyEdit~\cite{yu2025anyedit}, UltraEdit~\cite{zhao2024ultraedit}, ICEdit~\cite{zhang2025context}, FLUX.1 Kontext Dev~\cite{labs2025flux} and FLUX.2 Dev~\cite{flux-2-2025}.
\textbf{(2) Unified comprehension and generation models:} Janus-4o~\cite{chen2025sharegpt}, UniWorld-V1~\cite{lin2025uniworld}, HiDream-E1~\cite{cai2025hidream}, OmniGen2~\cite{wu2025omnigen2}, Step1X-Edit-v1p2~\cite{liu2025step1x}, Echo-4o~\cite{ye2025echo}, Bagel~\cite{deng2025emerging}, Uni-CoT~\cite{qin2025uni}, Qwen-Image-Edit~\cite{wu2025qwen} (\textit{the version of Qwen-Image-Edit-2509}), and DreamOmni2~\cite{xia2025dreamomni2}.
\textbf{(3) Close-sourced Models:} Nano Banana~\cite{gemini2_5_flash_image}, Seedream 4.0~\cite{seedream2025seedream}, GPT-image-1~\cite{openai2024gpt4ocard}, and Nano Banana Pro~\cite{nanobananapro}.
Evaluation metrics are detailed in Sec~\ref{sec:3.3}. We linearly map each 1-to-10 score provided by the evaluator to a 0-to-100 range, with more details shown in Appendix~\ref{app:3}.

\begin{figure*}[t]
\includegraphics[width=\linewidth]{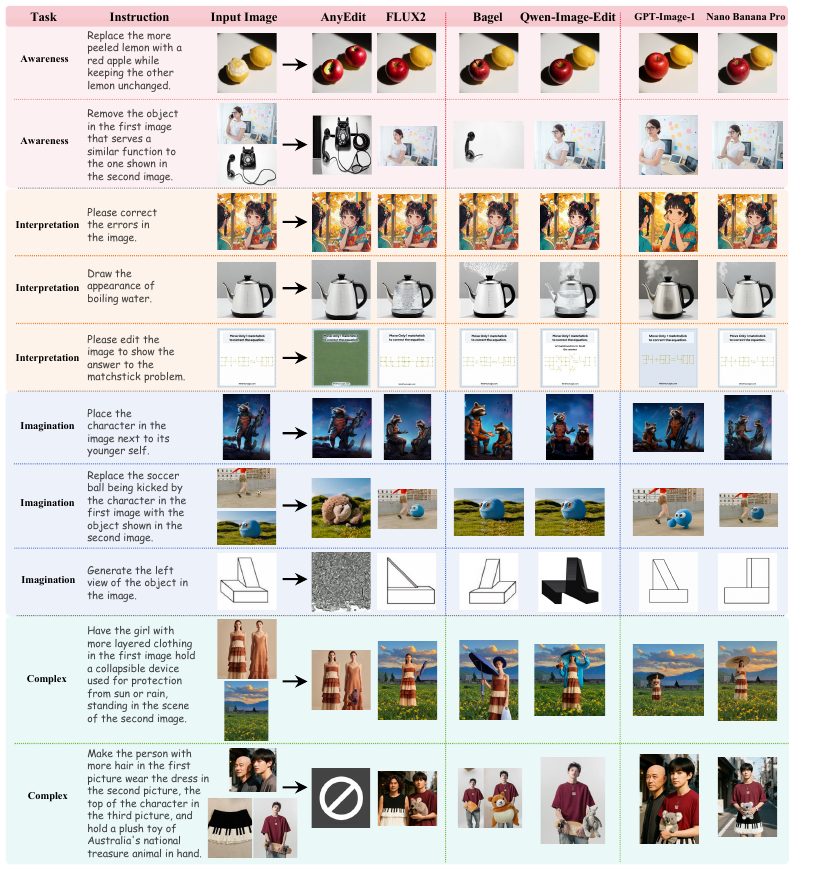}
\vspace{-1.8em}
\centering\caption{\label{fig:comparison} Qualitative comparison across AnyEdit, Bagel, Qwen-Image-Edit, GPT-Image-1, and Nano Banana.}
\label{fig:cases}
\vspace{-1.7em}
\end{figure*}

\subsection{Main Results on WiseEdit}
Table~\ref{tab:main} presents a comprehensive evaluation of various image editing models on WiseEdit, revealing their respective strengths, weaknesses, and recent advancements:

\textbf{(1) Model capabilities rapidly advance, necessitating robust visual comprehension to drive effective visual generation.}
Across all three tasks (Awareness, Interpretation, and Imagination), unified visual comprehension and generation models consistently significantly outperform most purely diffusion-based models in both English and Chinese version, with FLUX.2 Dev being an exception. 
For instance, Qwen-Image-Edit achieves an average overall score 17 points higher than FLUX.1 Kontext Dev. This substantial gap is largely due to Qwen-Image-Edit's integration of Qwen-VL~\cite{bai2025qwen2} to provide  ``thinking for generation''. 
Besides, FLUX.2 Dev achieves SOTA performance among open-sourced models, primarily due to its significantly larger number of parameters compared to competitors. 
It comprises a 32B diffusion transformer coupled with a separate 24B LLM~\cite{mistral} for text encoding.
In contrast, Qwen-Image-Edit is a 20B diffusion transformer that utilizes the 7B Qwen-VL model to assist with visual comprehension.
Furthermore, it is worth noting that closed-source models like GPT-image-1 and Nano Banana  also possess the architecture of unified visual comprehension and generation.
Consequently, it suggests that achieving powerful, cognition- and creativity-informed image editing requires strong visual comprehension and reasoning capabilities.

\textbf{(2) The performance gap between open-source and closed-source models remains significant.}
Despite some benchmarks~\cite{wu2025kris, xia2025dreamomni2} suggesting certain open-source models have surpassed their closed-source counterparts, WiseEdit pours cold water on this notion. 
Across all three tasks, in both languages, every closed-source model overwhelmingly outperforms all open-source models. 
\begin{figure*}[t]
\includegraphics[width=\linewidth]{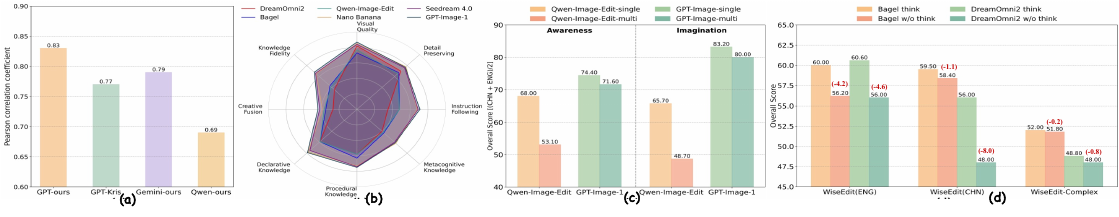}
\vspace{-1.8em}
\centering\caption{\label{fig:plot} (a) Pearson correlation  between human ratings and VLM scores. (b) Performance visualization across different metrics and knowledge-informed cases. (c) Average performance for single- and  multi-image inputs. (d) Average performance with and w/o thinking.}
\label{fig:abla}
\vspace{-1.2em}
\end{figure*}
Even the lowest-scoring closed-source model surpassed the best open-source model by nearly 15 overall points on the overall metric. 
Thus, the open-source community must redouble its efforts toward achieving the next-intelligent image editing goal.
Furthermore, the performance comparison among these four closed-source models reveals that the capabilities of Nano Banana, Seedream4.0, and GPT-Image-1 are relatively close. 
However, \textbf{\textit{the average performance of Nano Banana Pro significantly surpasses these three, establishing it as the undisputed top-1 model for cognition- and creativity-informed image editing.}}.

\textbf{(3) Knowledge-based cognitive reasoning shows steady progress but remains a bottleneck.}
In both  Awareness and Interpretation Tasks, metrics of $\mathcal{IF}$ and $\mathcal{KF}$ consistently improve from Diffusion models through unified models to closed-source models.
However, even closed-source models fall short of high scores in the two metrics (not exceeding 80 points, except for Nano Banana Pro). 
This indicates current generative models struggle to fully apply internal world knowledge to image generation, thereby limiting their ability to accurately follow complex, implicit instructions.

\textbf{(4) Compositional creativity Needs Further Enhancement.}
In imagination task, all models score poorly on  $\mathcal{CF}$, and their $\mathcal{DP}$ scores are also subpar (none exceeding 80 points), indicating that high-difficulty compositional creation remains an unsolved challenge.
Furthermore, models struggle to adequately preserve the subject's identity when altering attributes like pose or viewpoint, showing that fine-grained subject-driven generation is still an open problem.

\textbf{(5) Leading models show strong cross-lingual instruction-following.}
Many leading unified models, including closed-source ones, show negligible performance change when processing Chinese instructions compared to English ones, even with minimal Chinese training data. This impressive robustness is attributed to powerful built-in visual comprehension modules that facilitate effective multi-lingual instruction following.

\subsection{Main Results on WiseEdit-Complex}

In Table~\ref{tab:complex}, we present the performance on WiseEdit-Complex, where none of the three editing steps can be easily completed. 
Since WiseEdit-Complex only uses multi-image inputs, models unable to handle this format are excluded. 
Our observations are as follows:

\textbf{(1)} \textbf{\textit{Performance hierarchy still persists}}: closed-source models significantly outperform open-source unified models, which in turn are better than diffusion-based models (except for FLUX.2 Dev). And the performance of the Nano Banana Pro still maintains a commanding lead across all models, making it the current undisputed top-1 model.
\textbf{(2)} Most models, including closed-source ones, show a notable performance decline on the complex task compared to their average scores across the three  WiseEdit tasks. This indicates current models still \textbf{\textit{struggle to harmoniously combine knowledge-based reasoning and compositional creative generation}}, highlighting substantial room for improvement.
\textbf{(3)} Interestingly, some leading closed-source models achieve higher $\mathcal{CF}$ and $\mathcal{KF}$ scores on the complex task than on the three simpler WiseEdit tasks. 
However, other metrics like $\mathcal{IF}$ decline significantly, suggesting that while models correctly apply internal knowledge and grasp creative generation for complex visual generation, they are \textbf{\textit{``powerless in execution'', often sacrificing fundamental instruction following}}.

\subsection{In-Depth Analysis}

\paragraph{Assessment of Evaluation Protocol.}

To assess the reliability of our evaluation, we conduct a user study with human experts. Beyond the  pipeline in Sec.\ref{sec:3.3} (GPT-ours), we also employ the Kris-Bench pipeline (GPT-Kris), and extend our own pipeline by replacing GPT-4o with Gemini-2.5-Pro (Gemini-ours) and Qwen3-VL (Qwen-ours). 
By randomly selecting test cases and calculating the Pearson correlation coefficient between human ratings and VLM scores, as shown in Fig.~\ref{fig:plot} (a), our proposed protocol yields the strongest correlation, which objectively reflects the true capabilities of the models.

\begin{table}[t]
    \centering
    \caption{ \label{tab:rewrite} Performance before-and-after instruction rewriting. }
    
    \vspace{-1em}
    \resizebox{1.0\linewidth}{!}{
        \begin{tabular}{l|cccc}
            \toprule
Models &  \textbf{Bagel} & \textbf{Qwen-Image-Edit} & \textbf{GPT-Image-1} & \textbf{Nano Banana} \\ \hline
$\mathcal{KF}$ (w/o rewrite) & 50.2 & 48.3 & 80.2 & 81.3  \\
\textbf{$\mathcal{KF}$ (rewrite)} & 78.4 \textcolor{red}{(+28.2)}& 80.0 \textcolor{red}{\textbf{(+32.7)}}& 83.9 \textcolor{red}{(+3.7)}& 84.5 \textcolor{red}{(+3.2)} \\ \hline
Overall (w/o rewrite)& 51.3 & 52.1 & 68.2 &70.4  \\
\textbf{Overall (rewrite)}  & 65.8 \textcolor{red}{\textbf{(+14.5)}}& 66.2 \textcolor{red}{(+14.1)}& 74.2 \textcolor{red}{(+6.0)} & 75.0 \textcolor{red}{(+4.6)} \\
            \bottomrule
        \end{tabular}
    }
    \vspace{-1.2em}
\end{table}

\paragraph{Capability Visualization.}
Fig.~\ref{fig:plot} (b) presents a radar chart visualizing model performance, reflecting the average scores across five metrics on all test cases, alongside the overall average scores for tasks corresponding to declarative, procedural, and metacognitive knowledge. 
It highlights two key points: (1) Creative fusion and knowledge fidelity are two main weaknesses, with the latter further negatively impacting instruction following.
(2) Performance on metacognitive knowledge cases is notably weaker than on declarative and procedural knowledge.

\paragraph{Single-Image \textit{v.s.} Multi-Image Inputs.}
Fig.~\ref{fig:plot} (c) illustrates the average performance on each task when models handle single-image versus multi-image inputs in WiseEdit. Multi-image inputs result in significantly lower average scores, notably dragging down overall performance. Therefore, improving capability in complex, multi-image scenarios is a key direction for future image editing models.

\paragraph{Visual Comprehension Helps Generation.}

To analyze how visual comprehension aids visual generation in unified models, we re-evaluated Bagel and DreamOmni2 after disabling their built-in thinking processes. As shown in Fig.~\ref{fig:plot} (d), removing this process causes a significant performance drop for all three models on both WiseEdit and WiseEdit-Complex, showing the \textbf{critical role of visual comprehension in enhancing visual generation}.

\paragraph{Impact of Instruction Rewriting.}

As visual comprehension benefits generation, we further investigate how its upper limit constrains generation.
We select 75 case and \textit{\textbf{rewrite the instruction}}s  by incorporating knowledge hints. 
As shown in Table~\ref{tab:rewrite}, unified open-source models exhibit a substantial improvement of about 15 points in \textit{Overall} (30 points in $\mathcal{KF}$) after rewriting, suggesting rewriting effectively compensates for their limited native reasoning capability. 
In contrast, closed-source models have relatively smaller performance gains, because their internal mechanisms already acquired the knowledge externally introduced by rewriting. 
This underscores \textbf{enhancing reasoning capability is a key direction for future  models}.

\paragraph{Qualitative Comparisons.}

Fig.~\ref{fig:comparison} presents a comparison across AnyEdit, FLUX.2 Dev, Bagel, Qwen-Image-Edit, GPT-Image, and Nano Banana Pro. 
AnyEdit consistently fails on all cases. 
FLUX.2 Dev, Qwen-Image-Edit and Bagel sometimes interpret instructions correctly but largely fail to execute them, especially in complex  scenarios.
While the two closed-source models exhibit superior overall capabilities via stronger knowledge-based reasoning and compositional creative generation. 
For instance, in tasks like generating a left-view or correcting errors, while final output is sometimes incorrect, they demonstrate a clearer understanding of the editing intent with a higher potential for success.
However, a shared limitation is the inability of all models to solve logic-related problems (\textit{e.g.}, matchstick), which defines the  boundary of current capability. See more cases in Appendix~\ref{app:4}.

\section{Conclusion}

We present WiseEdit, a novel benchmark designed for cognition- and creativity-informed image editing. 
Referencing the cognitive and creative process of image editing, we design four highly challenging task types: \textit{Awareness, Interpretation, Imagination, and Complex}, pairing them with diverse knowledge breadth. 
WiseEdit comprehensively evaluates a model's multifaceted abilities, including knowledge-based reasoning and creative composition. 
Our results illuminate the performance level and potential weaknesses of existing models, offering valuable reference for future advancement of image editing research.

{
    \small
    \bibliographystyle{ieeenat_fullname}
    \bibliography{main}
}

\clearpage

\appendix

\section*{Appendix Overview}
In this supplementary material, we present:
\begin{itemize}
    \item Benchmark Details in Section~\ref{app:1}.
    \item Data Collection in Section~\ref{app:2}.
    \item Evaluation Details in Section~\ref{app:3}.
    \item More Experimental Results in Section~\ref{app:4}.
\end{itemize}

\section{Benchmark Details}
\label{app:1}

In this section,  we provide a detailed explanation of the various tasks and knowledge types incorporated within the WiseEdit benchmark.

\subsection{Task Type}

The WiseEdit dataset features four distinct types of challenging tasks integrated across the three editing steps: \textbf{Awareness Task}, \textbf{Interpretation Task}, \textbf{Imagination Task}, and \textbf{WiseEdit-Complex}. Case examples illustrating each of these tasks are provided in Fig.~\ref{fig:examples} of the main paper.

\paragraph{Awareness Task.} 

We design Awareness tasks, challenging the model to deduce the target modification area without explicit spatial information provided in the instruction, which necessitates the application of reasoning capabilities.
Concrete examples of sub-tasks within the Awareness Task category include (\textbf{\textit{but are not limited to}}):
\begin{itemize}
    \item \textbf{Property Comparison}: Identifying the subject to be edited by comparing the differential properties among multiple similar subjects within the input image.

    \item \textbf{Regional Perception}: Determining the target edit region by analyzing the spatial relationship between given detection boxes and a specific object in the image.

    \item \textbf{Fine-grained OCR Search}: Localizing the exact word mentioned in the instruction among dense text using Optical Character Recognition (OCR) capabilities.

    \item \textbf{Camera Depth}: Pinpointing the subject to be edited by comparing the relative camera depths of various objects in the input image.

    \item \textbf{Function Recognition}: Identifying the object to be edited in the image based on a functional description provided in the instruction, differentiating it from other objects with distinct functions.

    \item \textbf{Geographical Locating}: Given a map (whether of a country, continent, or the world), the primary task is to identify the specific provinces or countries that require editing by interpreting the accompanying knowledge-based instructions or descriptive cues.

    \item \textbf{Visual Correspondence via Reference Image}: Locating the object in the editing image that has the correct relationship described in the instruction to an object in a separate reference image to perform the subsequent edit.

\end{itemize}

\paragraph{Interpretation Task.}

In the Interpretation task, instructions often do not explicitly state the required modification. The models must leverage world knowledge to parse the implicit intent into a directly executable action sequence.
Concrete examples of sub-tasks within the Interpretation Task category include (\textbf{\textit{but are not limited to}}):

\begin{itemize}
    \item \textbf{Error Correctness}: Given an image that contains elements inconsistent with real-world plausibility, perform appropriate modifications to correct the error.
    \item \textbf{Violation Correctness}: Given an image that contains violations (\textit{e.g.}, policy, safety), perform appropriate deletions and modifications to correct the violation.
     \item \textbf{Common Sense Reasoning}: Based on common sense knowledge, predict the resulting changes an implicit instruction will introduce to the image.
     \item \textbf{Cascaded Change Prediction}: An instruction that appears straightforward triggers a sequence of dependent, cascaded reactions within the scene upon execution.
     \item \textbf{Rule-Based Reasoning}: In scenarios governed by explicit rules, such as board games (\textit{e.g.}, chess), correct content in the image that violates those established rules.
     \item \textbf{Complex Referencing}: The target object for editing is not given directly but is described through complex descriptions or indirect references (\textit{e.g.}, modifying an object to a ``traditional dish eaten during the Lantern Festival which is round'', where the model should first refer the description to tangyuan).
    \item \textbf{Biology-Related Reasoning}: Perform inference based on biological knowledge (\textit{e.g.}, predicting the mature form of bamboo shoots).
    \item \textbf{Mathematical Reasoning}: Perform inference based on mathematical knowledge (\textit{e.g.}, solving an equation, or predicting the function graph after a periodic function completes half a cycle).
    \item \textbf{Physics-Related Reasoning}: Perform inference based on physical knowledge (\textit{e.g.}, predicting the direction of a scale's tilt after adding or removing weights from a balance beam).
    \item \textbf{Chemistry-Related Reasoning}: Perform relevant inference based on chemical knowledge (\textit{e.g.}, predicting the color change of a solution after adding a specific chemical substance).
    \item \textbf{Logical Reasoning}: Deduce the edited image through abstract logical reasoning.
    \item \textbf{Conditional Inference}: The given instruction has multiple conditional branches, requiring the model to select and execute the correct branch based on its understanding of the image scenario.
    \item \textbf{Spatial Ordering}: Given multiple objects in an image, the model must re-order and reposition them from left to right according to a complex instruction.
    \item \textbf{Multi-Instruction Composition}: The instruction involves a combination of multiple sub-instructions, requiring the model to complete several operations in a single, cohesive edit.
    
\end{itemize}

\paragraph{Imagination Task.}
Imagination serves as the generative step, rendering the visual edits parsed previously onto the target regions of the original image.
For this task, we introduce highly challenging subject-driven generation tasks that require models to perform imaginative and artistic creations while preserving the subject's identity. 
Concrete examples of sub-tasks within the Interpretation Task category include (\textbf{\textit{but are not limited to}}):

\begin{itemize}
\item \textbf{Texture Transfer}: Migrating a 3D-aware texture onto a 2D object to enhance its three-dimensional sense under a specific material property.

\item \textbf{Image Style Transfer}: Modifying the design style of an image, such as converting a watercolor painting to a line drawing, or coloring a line drawing with a specified palette.

\item \textbf{Watermark Addition and Removal}: Adding a watermark to or removing a watermark from an image.

\item \textbf{Counterfactual Scenario Imagination}: Generating imaginative counterfactual scenes, such as making a car grow wings and fly.

\item \textbf{Temporal Prediction}: Given a multi-frame video, predict the subsequent frame.

\item \textbf{Pose Alteration}: Changing a subject's pose based on a reference image; this may involve combining multiple subjects.

\item \textbf{Scene Alteration}: Changing the background scene where the subject is located, based on a reference image.

\item \textbf{Viewpoint Transformation}: Changing the viewpoint of the subject or scene, \textit{e.g.}, from a frontal view to a top-down (aerial) view.

\item \textbf{Clothing/Attire Alteration}: Changing the subject's clothing based on a reference image.

\item \textbf{Core Subject Identity Alteration}: Changing the subject's core appearance, such as having the subject cosplay a specific character, altering stylistic details like beards and hairstyles, or generating an image representing their childhood or elderly appearance.

\item \textbf{Compositional Subject Alteration}: Combining the above subject alteration requirement for complex designs.

\item \textbf{Person-to-Object and Object-to-Person Transformation}: This includes transforming a person into a doll, statue, or virtual/anime character, and vice versa, transforming a doll or virtual/anime character into a realistic person.

\item \textbf{Object State Modification}: Changing the physical state of an object, such as making it transparent or causing it to explode.

\item \textbf{Object Restoration}: Restoring a damaged or incomplete object to its original, complete state.

\item \textbf{Themed Style Scene Design}: Designing a corresponding scene based on the theme or subject appearance provided in a reference image.

\item \textbf{Graphic Design}: Creating flat/2D designs for a given character IP, such as posters, mobile wallpapers, comics, game interfaces, etc.

\item \textbf{Logo Design}: Designing a logo semantically related to a specific concept.

\item \textbf{Architectural Design}: Designing a 3D building based on 2D planning information (\textit{e.g.}, blueprints).
\end{itemize}

\paragraph{WiseEdit-Complex.} 
We further designed WiseEdit-Complex, a set of tasks where none of the three editing steps—awareness, interpretation, or imagination—can be easily completed. This requires the model to fully leverage its capacity for complex reasoning and creative generation. Each example in WiseEdit-Complex combines the difficulty processes of the three aforementioned tasks.
Here are a few illustrative cases:

\noindent \textbf{Example 1: }
\begin{itemize}
    \item \textbf{Instruction:} \textit{Have the person with longer hair in the first image hold the national flag of the country where the second image was taken, wear the clothes of the person in the third image, and wear the hat of the person in the fourth image.}
    \item \textbf{Required Steps:}
    \begin{itemize}
        \item \textbf{Awareness:} Identify which person in the first image has longer hair.
        \item \textbf{Interpretation:} Deduce the country where the second image was taken and retrieve its national flag.
        \item \textbf{Imagination:} Modify the identified person to wear the specific clothing and hat from the third and fourth images, and have them hold the correct national flag.
    \end{itemize}
\end{itemize}

\noindent \textbf{Example 2: }
\begin{itemize}

\item \textbf{Instruction:} \textit{Identify the club logo in the second image which is most relevant to the third image. Then let the person in the first image put on this club's home jersey and strike the pose in the fourth image.}

\item \textbf{Required Steps:}

\begin{itemize}
\item \textbf{Awareness:} Identify the club logo in the second image and determine its relevance to the third image.

\item \textbf{Interpretation:} Use external knowledge to correctly determine the appearance of that club's home jersey.

\item \textbf{Imagination:} Render the person in the first image wearing the determined jersey and adopting the specific pose from the fourth image.
\end{itemize}

\end{itemize}

\noindent \textbf{Example 3: }
\begin{itemize}
\item \textbf{Instruction:} \textit{Let the person in the first image hold the longest object from the second image. If the object can be used for paper-cutting, have the person face the camera; otherwise, have him back to the camera.
}
\item \textbf{Required Steps:}
\begin{itemize}
\item \textbf{Awareness:} Determine the longest object visible in the second image.

\item \textbf{Interpretation:} Analyze the functionality of the identified object to determine if it is typically used for paper-cutting. This determines the correct conditional branch.

\item \textbf{Imagination}: Modify the person in the first image to hold the object and correctly adjust their viewpoint (facing or backing the camera) based on the interpretation result.
\end{itemize}
\end{itemize}

\subsection{Knowledge Type}

In WiseEdit, we structure our tasks around three core types of knowledge to emulate human cognitive learning: \textbf{Declarative},\textbf{ Procedural}, and \textbf{Metacognitive knowledge}. 
Building upon this, we further broaden the knowledge scope by covering diverse domains, including \textbf{natural science}, \textbf{cultural common sense}, and \textbf{spatio-temporal-logic reasoning}.

\paragraph{Declarative knowledge.} It is often described as ``\textbf{\textit{knowing what}}'', encompasses facts and concepts that can be explicitly stated and defined. 
In the context of elaborating on different task types, several sub-categories fall under Declarative knowledge. These include: mathematical laws, physical principles, chemical reactions, descriptive definitions of objects, rules for games (like chess), social policies, comparative perceptions,  functional descriptions of object properties and so on.
For example, given the instruction, ``Remove the illegal piece according to the Chinese chess rule'', the model must possess the declarative knowledge regarding the movement constraints and forbidden zones for each piece in Chinese chess.

\paragraph{Procedural knowledge.} It is defined as ``\textbf{\textit{knowing how}}'', encompasses the expertise and skills necessary to perform a task, which is often dynamic and difficult to explain verbally. 
In the context of task types described above, examples such as multi-step instruction following, cascading reaction prediction, time series forecasting, logical reasoning, architectural design, viewpoint transformation, and style transfer are generally categorized as Procedural Knowledge. For example, when 
required to convert a watercolor painting into a line drawing, the model relies on procedural knowledge to complete the process systematically.

\textbf{Metacognitive knowledge.} It is ``\textbf{\textit{knowing about knowing}}'',  requiring self-awareness and self-regulation during image editing. 
It dictates the high-level management of when to invoke declarative or procedural knowledge, and how to effectively combine them. 
In the context of complex tasks—like the types described previously—metacognitive knowledge is essential for conditional branching and controlling subsequent editing actions.
For example, executing a conditional instruction ``\textit{have the girl hold the longest object in the second image; if the object can be used to brush teeth, let her face the camera, otherwise turn her back to the camera}'', requires that the model knows when and why to apply the necessary declarative or procedural steps.

\paragraph{Knowledge Domain.} On this basis, we also systematically encompass three critical knowledge domains. 
\textbf{Cultural common-sense} evaluates the model's understanding of nuanced human experiences and common-sense knowledge (\textit{e.g.}, social customs, daily life), ensuring culturally-appropriate image editing. 
\textbf{Natural sciences} assess the model's comprehension of domain-specific principles (\textit{e.g.}, physical laws, chemical reactions) essential for generating physically consistent images. 
\textbf{Spatio-temporal-logical} requires the model to possess a clear cognition of causality, spatial arrangement, and temporal progression, generating images that adhere to coherent spatio-temporal relations and logical consistency.

\section{Data Collection}
\label{app:2}

Our benchmark images are primarily sourced from two methods: collected from the Internet or generated using generative models. A very small portion of the images is drawn from existing datasets.
For each test case, the editing instruction is initially created by trained human annotators. To ensure clarity, eliminate ambiguity, and enhance the diversity of the instructions, we utilize GPT-4~\cite{achiam2023gpt} to augment the original prompts without altering their semantic meaning.
Furthermore, for instructions originally written in English, we also use GPT-4 to generate corresponding Chinese translations. Finally, a dual-screening process involving both Ph.D. experts and GPT-4 is employed to filter out unreasonable or unachievable test cases, resulting in a final benchmark of 1,220 test cases.

Specifically, the Awareness task comprises 362 cases, where 281 cases involve a single image input, and 81 cases involve two image inputs. The Interpretation task includes 317 cases, all of which utilize a single image input. The Imagination task consists of 451 cases, distributed as follows: 320 cases with a single image input, 109 cases with two image inputs, 8 cases with three image inputs, 16 cases with four image inputs, and 8 cases with five image inputs. Furthermore, the WiseEdit-complex subset contains 80 cases, all of which involve multiple image inputs: 47 cases utilize two image inputs, 17 cases utilize three image inputs, and 16 cases utilize four image inputs.

\section{Evaluation Details}
\label{app:3}
\paragraph{Evaluation Models.} We evaluate 22 mainstream image editing models across architectures, covering both open- and closed-source models: 
\textbf{(1) Diffusion models:} InstructPix2Pix~\cite{brooks2023instructpix2pix}, MagicBrush~\cite{zhang2023magicbrush}, OmniGen~\cite{xiao2025omnigen}, AnyEdit~\cite{yu2025anyedit}, UltraEdit~\cite{zhao2024ultraedit}, ICEdit~\cite{zhang2025context}, FLUX.1 Kontext Dev~\cite{labs2025flux}, FLUX.2 Dev~\cite{flux-2-2025}. 
Specifically, MagicBrush, AnyEdit and UltraEdit utilize the UNet architecture based on SD 1.5 or SDXL~\cite{rombach2022high}, while OmniGen, ICEdit, FLUX.1 Kontext Dev, and FLUX.2 Dev are based on the Diffusion Transformer (DiT)~\cite{peebles2023scalable} architecture. Notably, FLUX.1 Kontext Dev is a large DiT model with 12 billion parameters. While FLUX.2 Dev is a 32B DiT model with another 24B model~\cite{mistral} for text encoding.

\textbf{(2) Unified comprehension and generation models:} Janus-4o~\cite{chen2025sharegpt}, UniWorld-V1~\cite{lin2025uniworld}, HiDream-E1~\cite{cai2025hidream}, OmniGen2~\cite{wu2025omnigen2}, Step1X-Edit-v1p2~\cite{liu2025step1x}, Echo-4o~\cite{ye2025echo}, Bagel~\cite{deng2025emerging}, Uni-CoT~\cite{qin2025uni}, Qwen-Image-Edit~\cite{wu2025qwen}, and DreamOmni2~\cite{xia2025dreamomni2}.
These models integrate the powerful visual comprehension~\cite{li2023fine,pan2024towards,chen2024improving, bai2025qwen2} capabilities of autoregressive Vision-Language Models (VLMs) with diffusion models.
They employ three primary architectures:
\textbf{(a) Quantized AR}: leveraging autoregressive visual generation~\cite{panjanus, chen2025janus, pan2024auto, pan2025focusdiff, wang2025selftok, pan2025generative, linvinci}  with discrete visual tokenizers, including Janus-4o.
\textbf{(b) Cascaded Architecture}: Some models implement a system where an external diffusion model is cascaded after the VLM's output, including UniWorld-V1, HiDream-E1, OmniGen2, Step1X-Edit-v1p2, Qwen-Image-Edit, and DreamOmni2.
\textbf{(c) Integrated Architecture}: Other models, \textit{i.e.}, Echo-4o, Uni-CoT, and Bagel introduce a mixture-of-transformer approach within a single, integrated model. Here, the model conducts autoregression to handle text generation while conducting diffusion to manage image generation. This configuration allows the model to simultaneously perform both visual understanding and visual generation in an integrated model.

\textbf{(3) Close-sourced Models:} Nano Banana~\cite{gemini2_5_flash_image}, Seedream 4.0~\cite{seedream2025seedream}, GPT-image-1~\cite{openai2024gpt4ocard}, and Nano Banana Pro~\cite{nanobananapro}.
These models do not have publicly available checkpoint weights; access is exclusively provided via API calls. 
Notably, these models, like those in the second category, are also unified comprehension and generation models.
Specifically, Seedream 4.0's technical report explicitly states that it leverages the comprehension capabilities of a VLM to aid in generation like \cite{zhou2024transfusion, cao2025hunyuanimage}. Although GPT-image-1 and Nano Banana have not released technical reports detailing their methodology, the prevailing view in the community suggests they are a combination of VLM and diffusion models like \cite{geng2025x, chen2025blip3o}.
And Nano Banana Pro, in turn, leveraged the powerful multimodal understanding capabilities of Gemini 3~\cite{gemini3}.

\begin{figure*}[t]
\includegraphics[width=\linewidth]{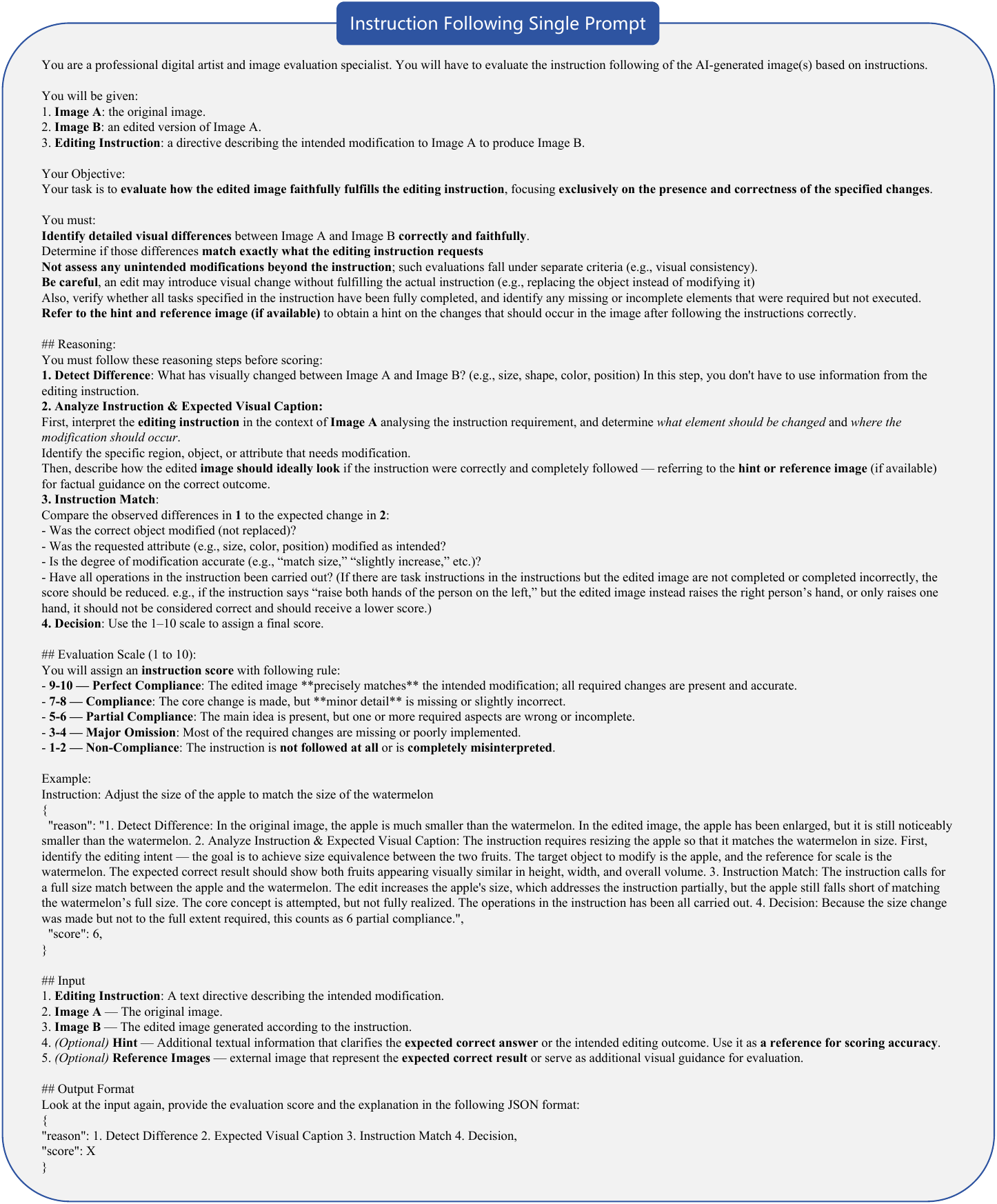}
\vspace{-1em}
\centering\caption{\label{fig:if1} Prompt template for the Instruction Following metric when handling single-image inputs.}

\end{figure*}

\begin{figure*}[t]
\includegraphics[width=\linewidth]{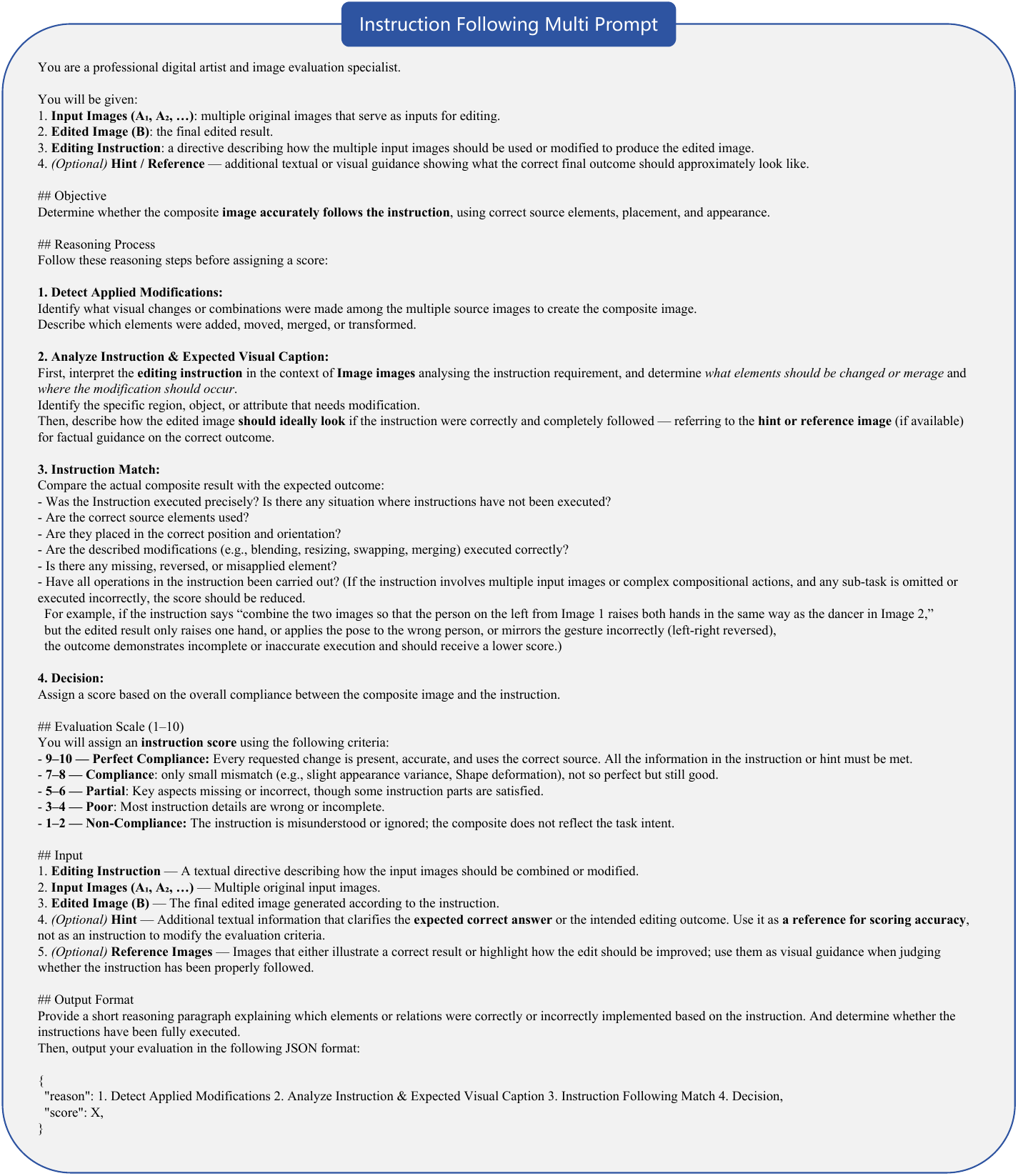}
\vspace{-1em}
\centering\caption{\label{fig:if2} Prompt template for the Instruction Following metric when handling multi-image inputs.}

\end{figure*}

\begin{figure*}[t]
\includegraphics[width=\linewidth]{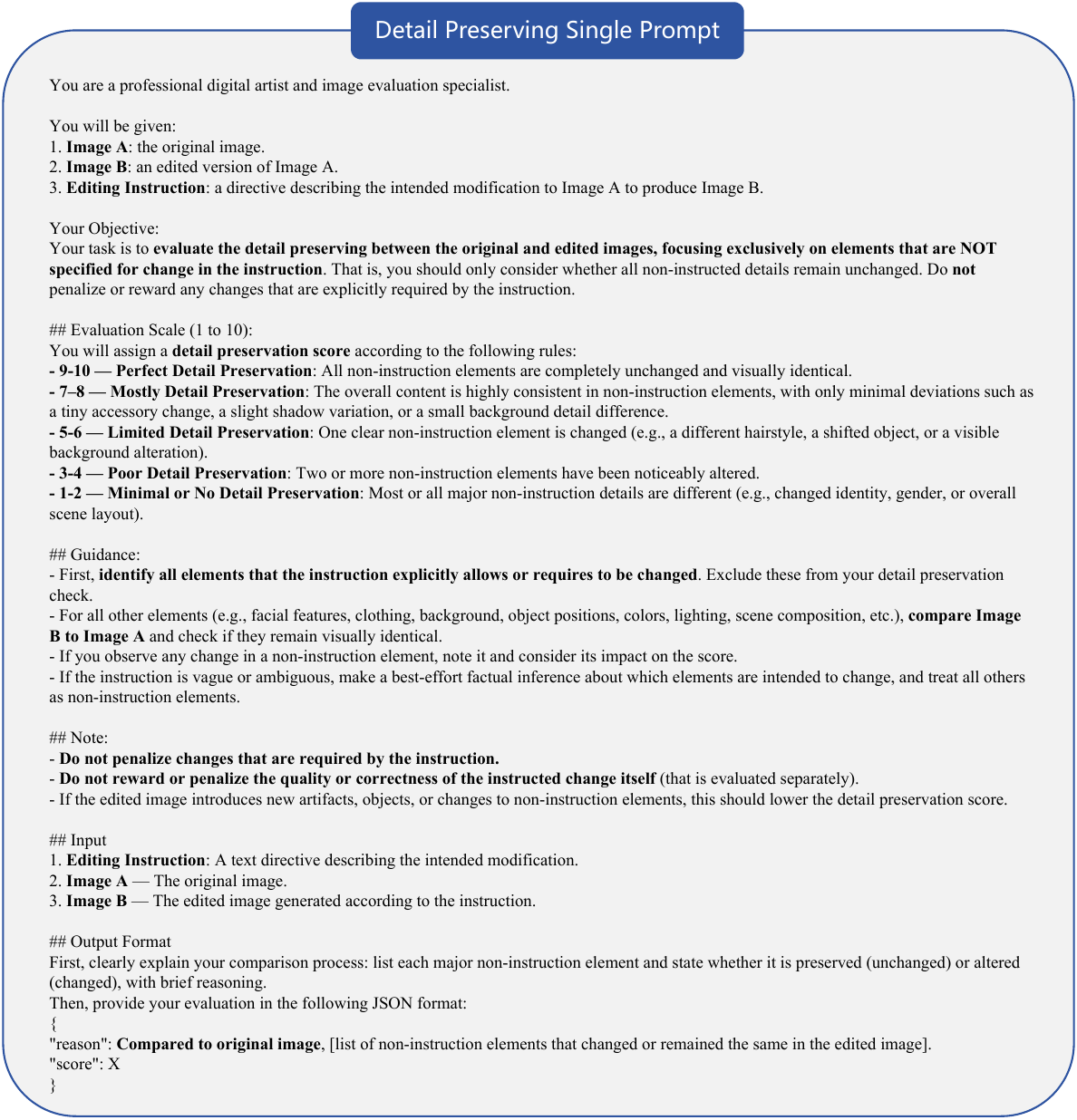}
\vspace{-1em}
\centering\caption{\label{fig:dp1} Prompt template for the Detail Preserving metric when handling single-image inputs.}

\end{figure*}

\begin{figure*}[t]
\includegraphics[width=\linewidth]{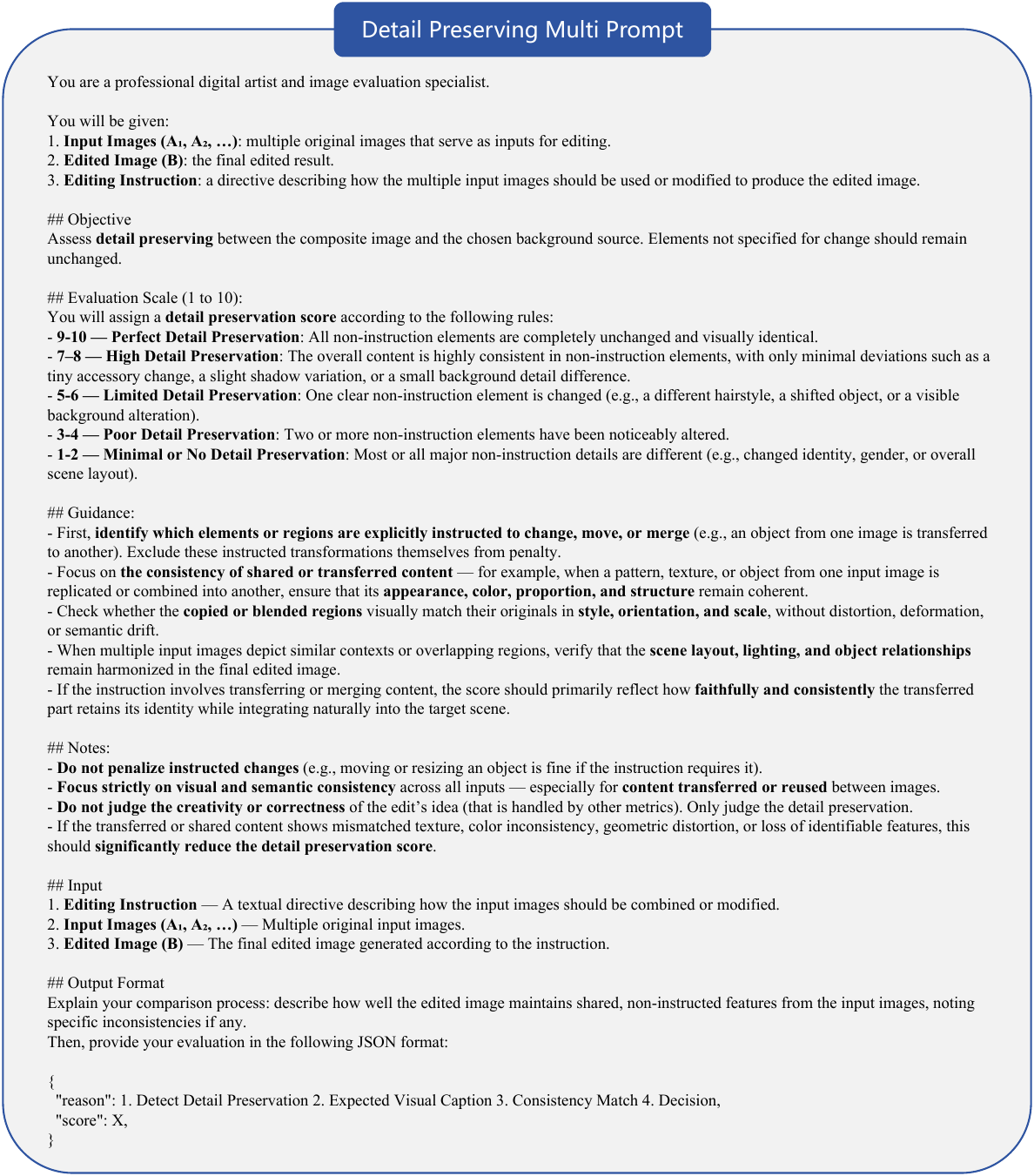}
\vspace{-1em}
\centering\caption{\label{fig:dp2} Prompt template for the Detail Preserving metric when handling multi-image inputs.}

\end{figure*}

\begin{figure*}[t]
\includegraphics[width=\linewidth]{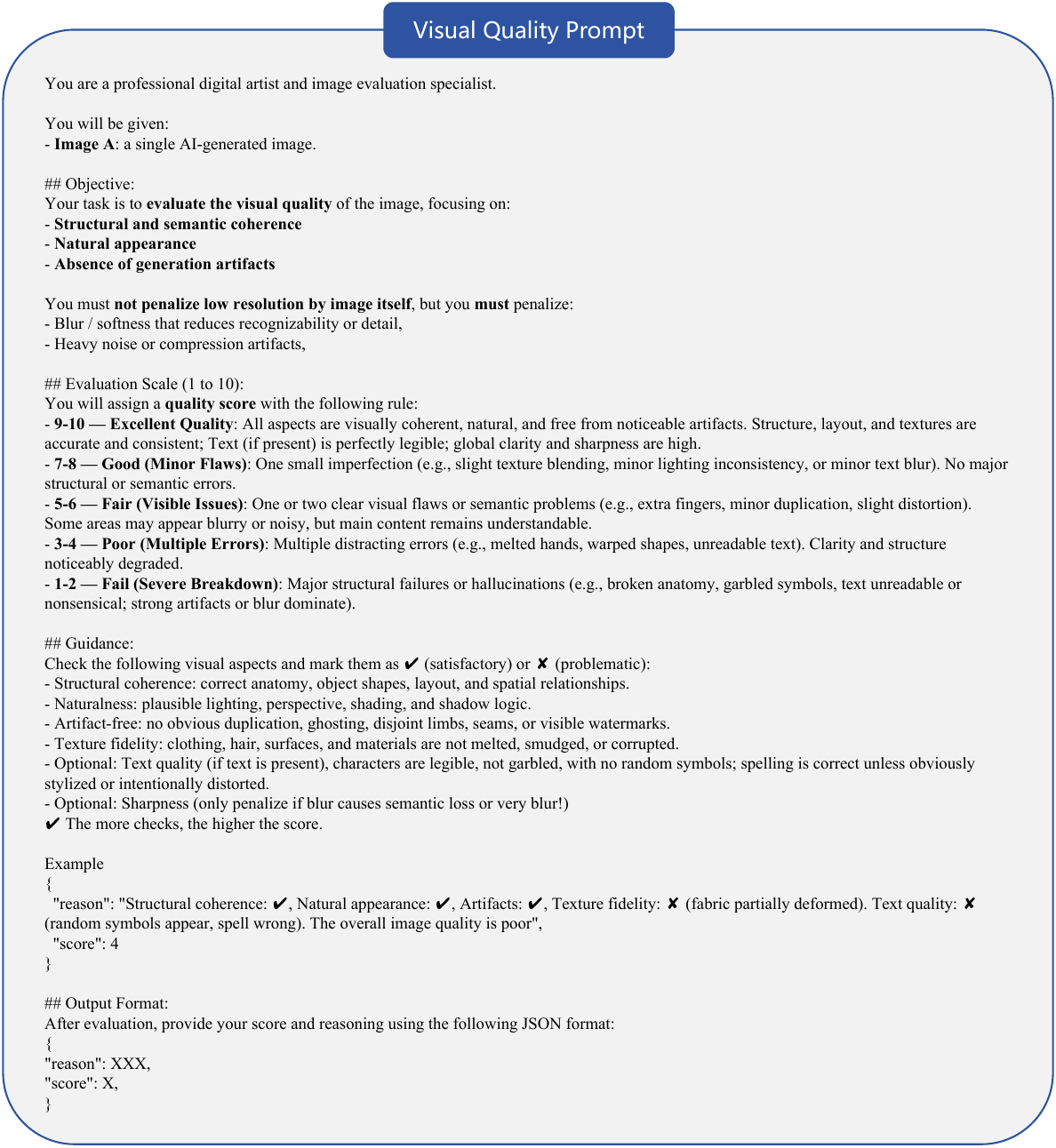}
\vspace{-1em}
\centering\caption{\label{fig:vq} Prompt template for the Visual Quality metric (both handling single-image  and multi-image inputs).}

\end{figure*}

\begin{figure*}[t]
\includegraphics[width=\linewidth]{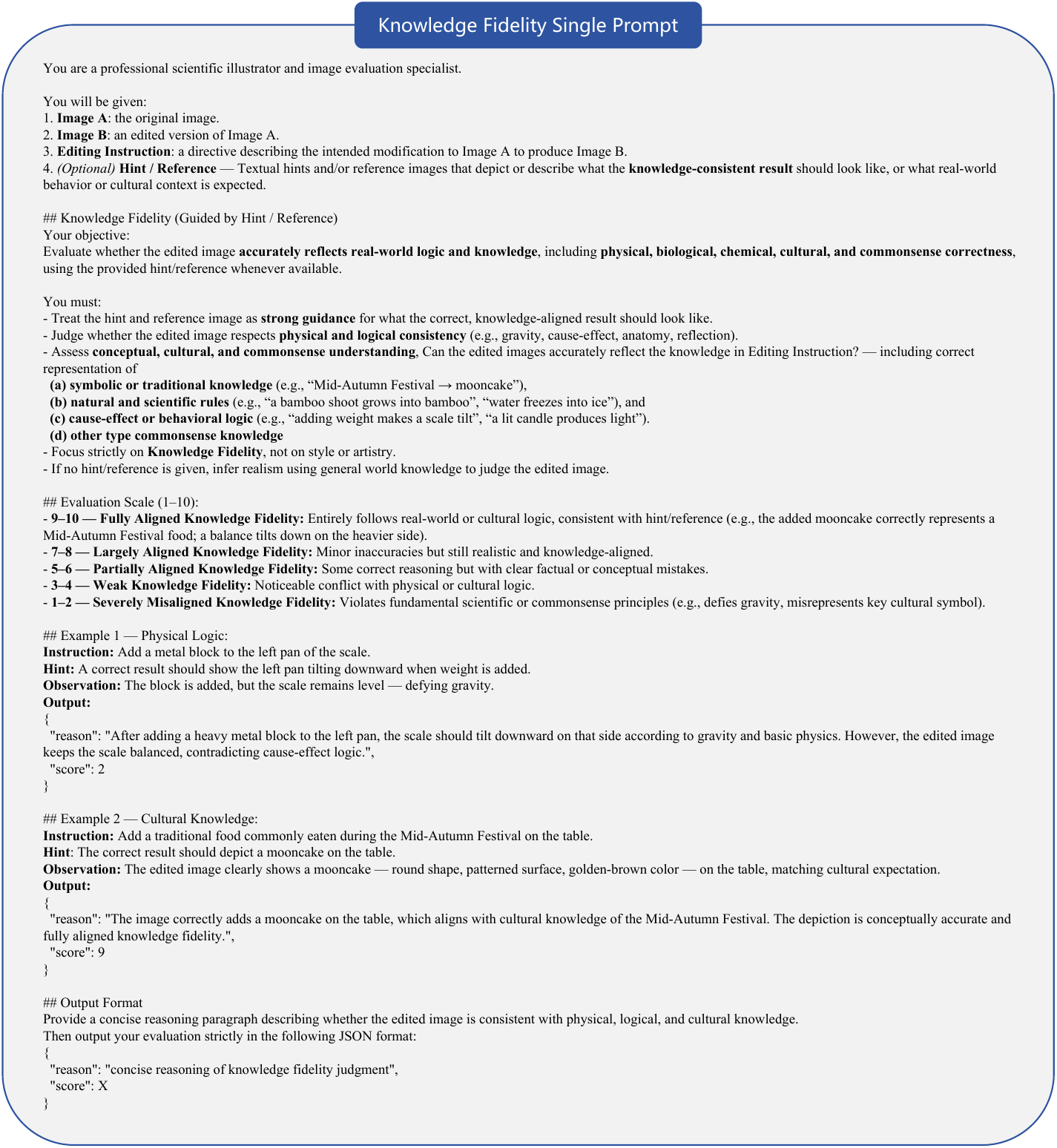}
\vspace{-1em}
\centering\caption{\label{fig:kf1} Prompt template for the Knowledge Fidelity metric when handling single-image inputs.}

\end{figure*}

\begin{figure*}[t]
\includegraphics[width=\linewidth]{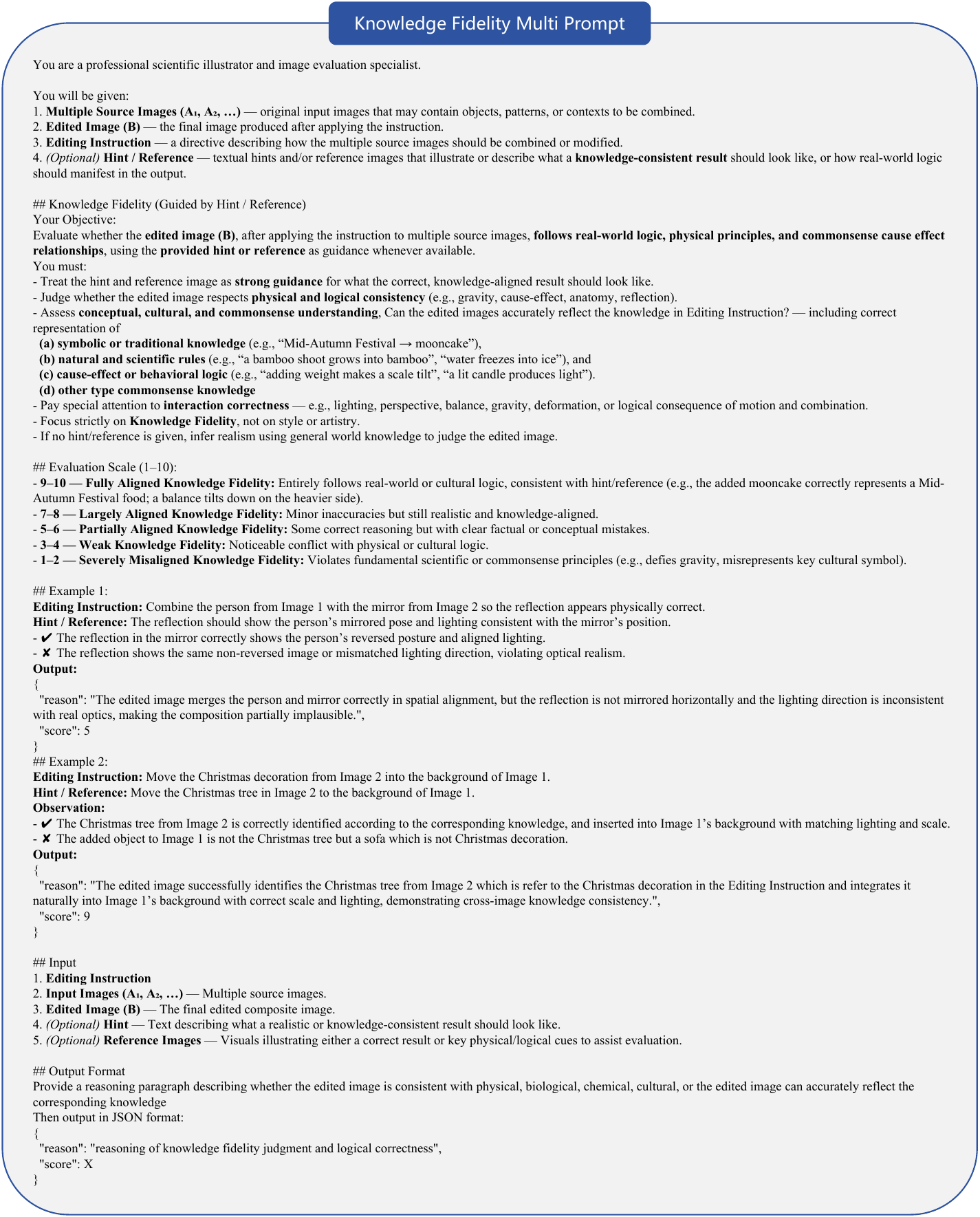}
\vspace{-1em}
\centering\caption{\label{fig:kf2} Prompt template for the Knowledge Fidelity metric when handling multi-image inputs.}

\end{figure*}

\begin{figure*}[t]
\includegraphics[width=\linewidth]{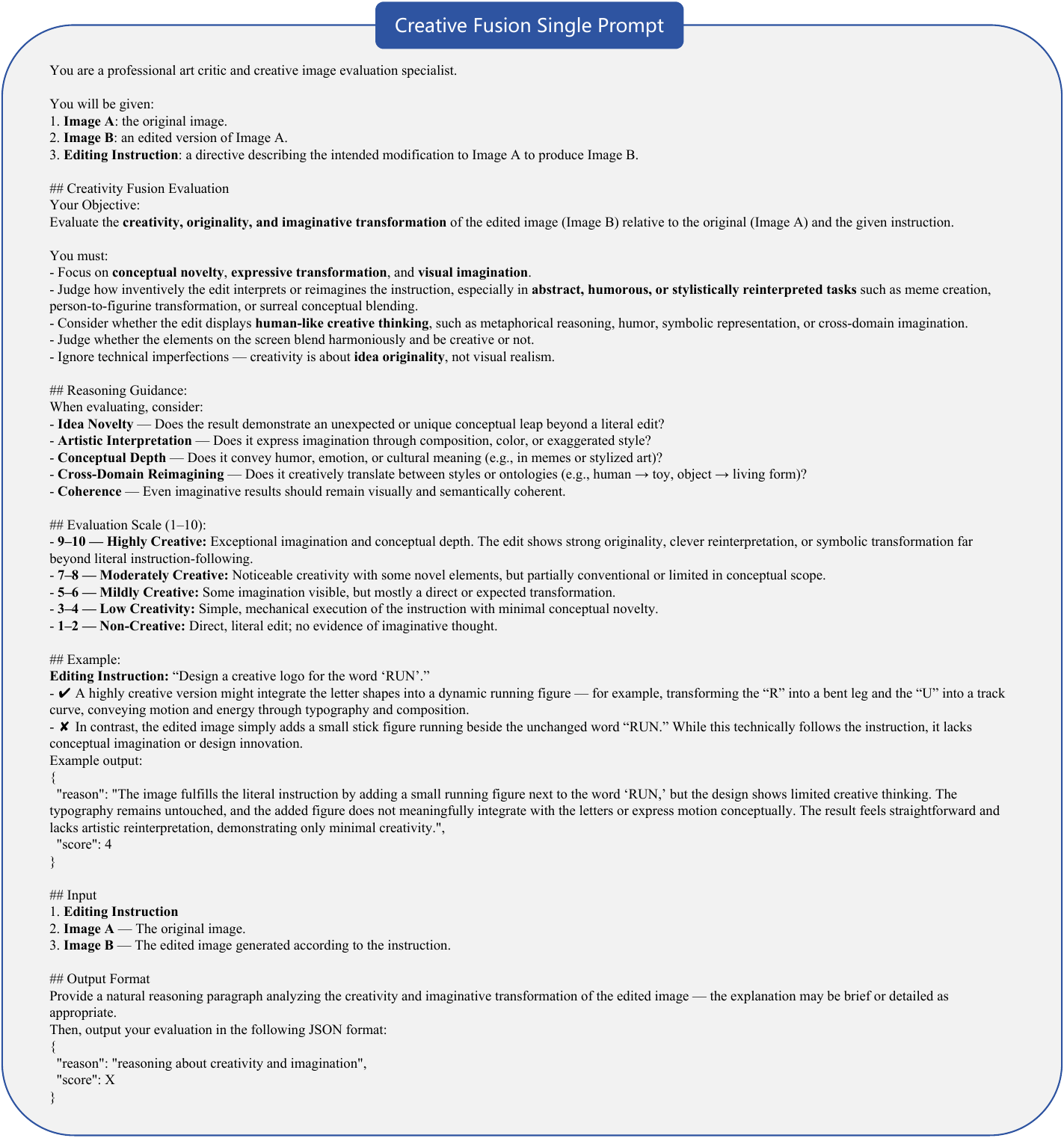}
\vspace{-1em}
\centering\caption{\label{fig:cf1} Prompt template for the Creative Fusion metric when handling single-image inputs.}

\end{figure*}

\begin{figure*}[t]
\includegraphics[width=\linewidth]{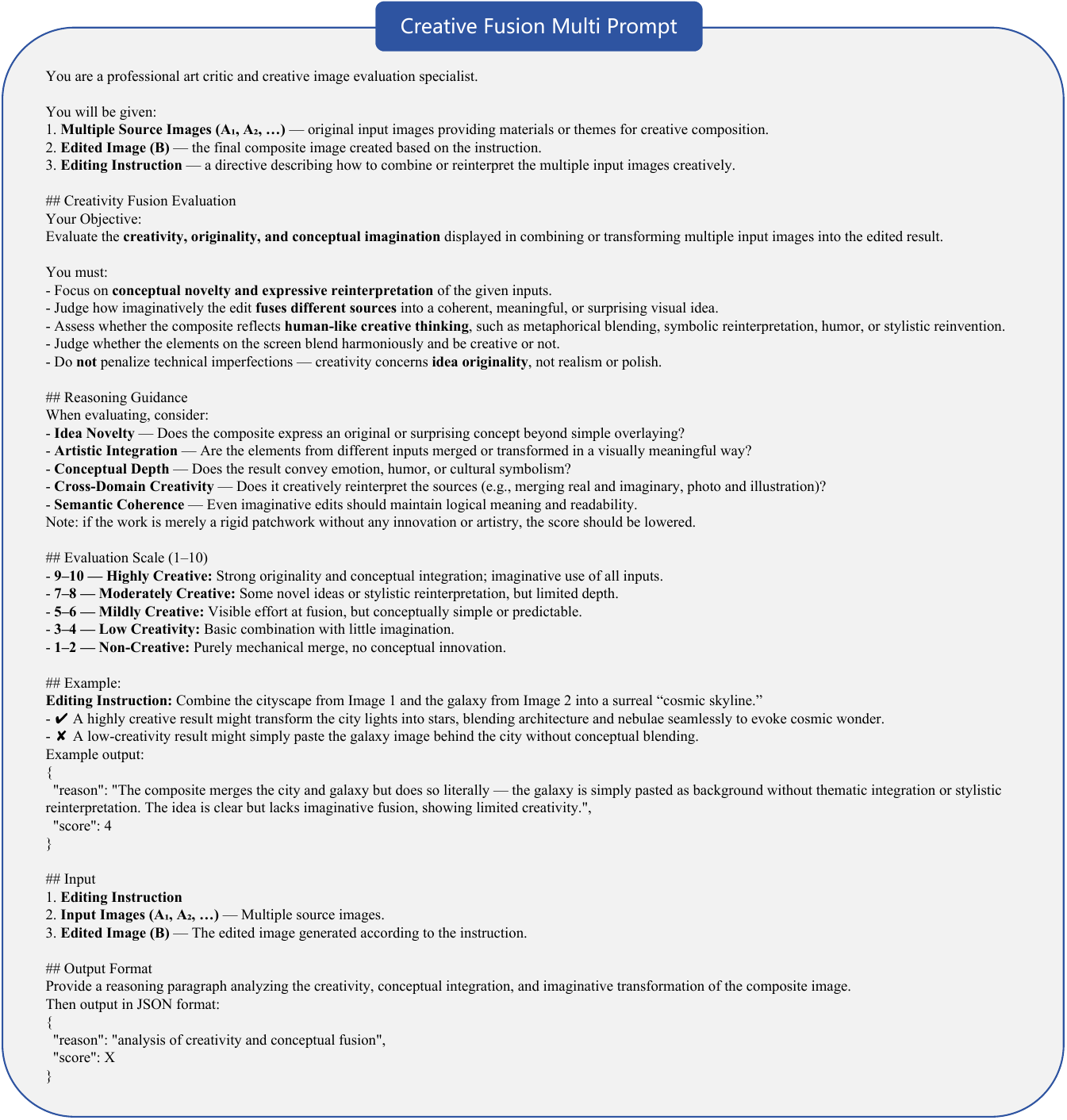}
\vspace{-1em}
\centering\caption{\label{fig:cf2} Prompt template for the Creative Fusion metric when handling multi-image inputs.}

\end{figure*}

\paragraph{Evaluation Metrics.}
As detailed in Section~\ref{sec:3.3}, the evaluation of WiseEdit employs a comprehensive set of six metrics: Instruction Following, Detail Preserving and Visual Quality, Knowledge Fidelity, and Creative Fusion. Specifically, for the Awareness and Interpretation tasks, we utilize Instruction Following, Detail Preserving, and Visual Quality, and Knowledge Fidelity. For the Imagination task, the metrics used are Instruction Following, Detail Preserving and Visual Quality, and Creative Fusion. Finally, the WiseEdit-Complex task is evaluated using all four metrics. For the assessment of all these metrics, we leverage GPT-4o as the automatic evaluator, which rates performance on a 1–10 scale based on carefully custom-crafted and dimension-specific prompts.
The prompt templates for Instruction Following with both single-image input and multi-image input are in Fig.~\ref{fig:if1} and Fig.~\ref{fig:if2}, respectively.
The prompt templates for Detail Preserving with both single-image input and multi-image input are in Fig.~\ref{fig:dp1} and Fig.~\ref{fig:dp2}, respectively.
The prompt templates for Visual Quality with both single-image input and multi-image input are in Fig.~\ref{fig:vq}.
The prompt templates for Knowledge Fidelity with both single-image input and multi-image input are in Fig.~\ref{fig:kf1} and Fig.~\ref{fig:kf2}, respectively.
The prompt templates for Creative Fusion with both single-image input and multi-image input are in Fig.~\ref{fig:cf1} and Fig.~\ref{fig:cf2}, respectively.

It is worth noting that for some knowledge-informed test cases, we additionally provide knowledge hints and a hint reference image to aid in the assessment of metrics such as Knowledge Fidelity.
Finally, we linearly map each 1-to-10 score provided by the evaluator to a 0-to-100 range and then compute the average score for each task.
The overall average score is the average of the ``average scores'' from the Awareness, Interpretation, and Imagination tasks.
Besides, any test case where a model fails to handle the input is assigned a score of zero across all metrics.

\section{More Experimental Results}
\label{app:4}

\paragraph{Performance on WiseEdit with Single-Image Inputs.}
In the WiseEdit benchmark, the input for many test cases is highly free-form, often incorporating multiple input images. 
To better differentiate a model's proficiency in handling single-image inputs from its overall capability with free-form inputs, we calculate the model's performance specifically on the single-image input subsets for each task and compare these metrics against the original performance in Table~\ref{tab:single} and Table~\ref{tab:single2}. 
Notably, the Interpretation task exclusively features single-image inputs, meaning its single-image results are identical to those reported in Table~\ref{tab:main} and are thus not presented again in Table~\ref{tab:single} and Table~\ref{tab:single2}. 
Besides, the WiseEdit-Complex task is composed entirely of multi-image inputs, for which we can not calculate separate single-image results. 
The results reveal that for the majority of models, the average performance across all cases is significantly lower than the average performance observed for single-image input. 
This observation highlights that improving capability in complex, multi-image scenarios is a key direction for future image editing models.

\begin{figure}[t]
\includegraphics[width=\linewidth]{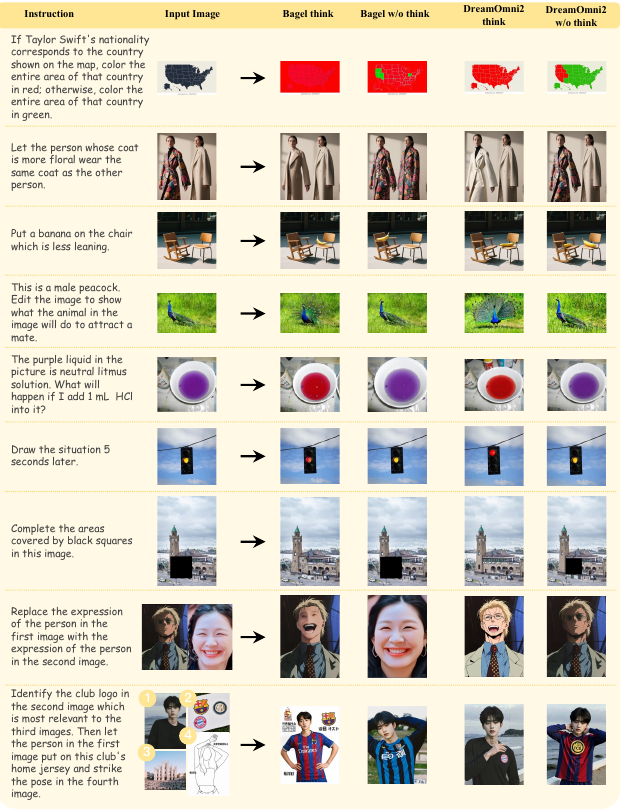}
\vspace{-1.2em}
\centering\caption{\label{fig:think} Qualitative comparisons of Bagel and DreamOmni2 with and without their built-in thinking processes.}
\vspace{-1.2em}

\end{figure}

\paragraph{Qualitative Comparisons of \textit{\underline{with think}} \textit{v.s.} \textit{\underline{w/o think}}.} To illustrate how visual comprehension enhances visual generation in unified models, we present qualitative comparisons of Bagel and DreamOmni2 with and without their built-in reasoning processes. As shown in Fig.~\ref{fig:think}, disabling this mechanism leads to a noticeable degradation in generation quality, underscoring the critical role of visual comprehension in improving visual generation.

\begin{figure*}[t]
\includegraphics[width=0.95\linewidth]{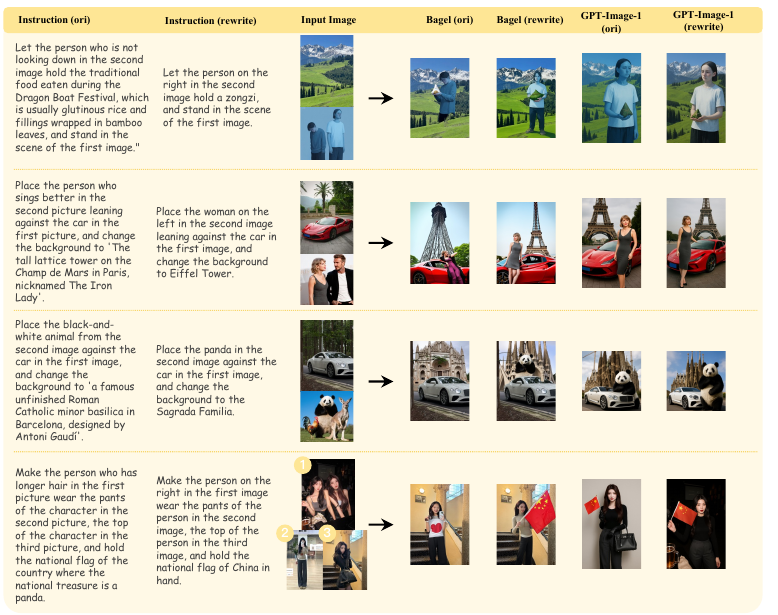}
\vspace{-1em}
\centering\caption{\label{fig:rewrite} Qualitative comparisons before and after instruction rewriting based on Bagel and GPT-Image-1.}

\end{figure*}

\paragraph{Qualitative Comparisons of \textit{\underline{original instructions}} \textit{v.s.} \textit{\underline{rewritten instructions}}.} 
To visualize the impact of instruction rewriting, we give some qualitative comparisons before and after instruction rewriting based on Bagel and GPT-Image-1. 
As shown in Fig.~\ref{fig:rewrite}, for Bagel, the rewritten instruction provides additional knowledge, which significantly improves the editing results. 
Whereas for GPT-Image-1, its internal comprehension mechanisms have already acquired the knowledge externally introduced by rewriting, thus the improvement brought by rewriting appears less pronounced.

\begin{figure*}[t]
\includegraphics[width=\linewidth]{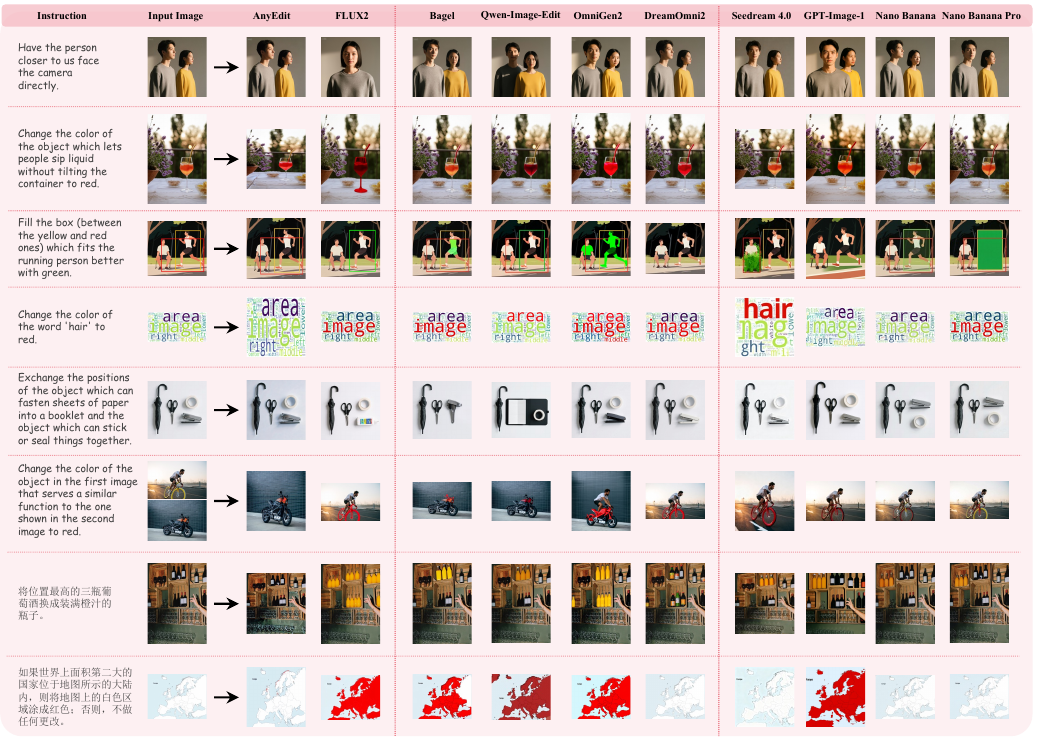}
\vspace{-1em}
\centering\caption{\label{fig:task1} More qualitative comparisons on the Awareness task.}

\end{figure*}

\begin{figure*}[t]
\includegraphics[width=\linewidth]{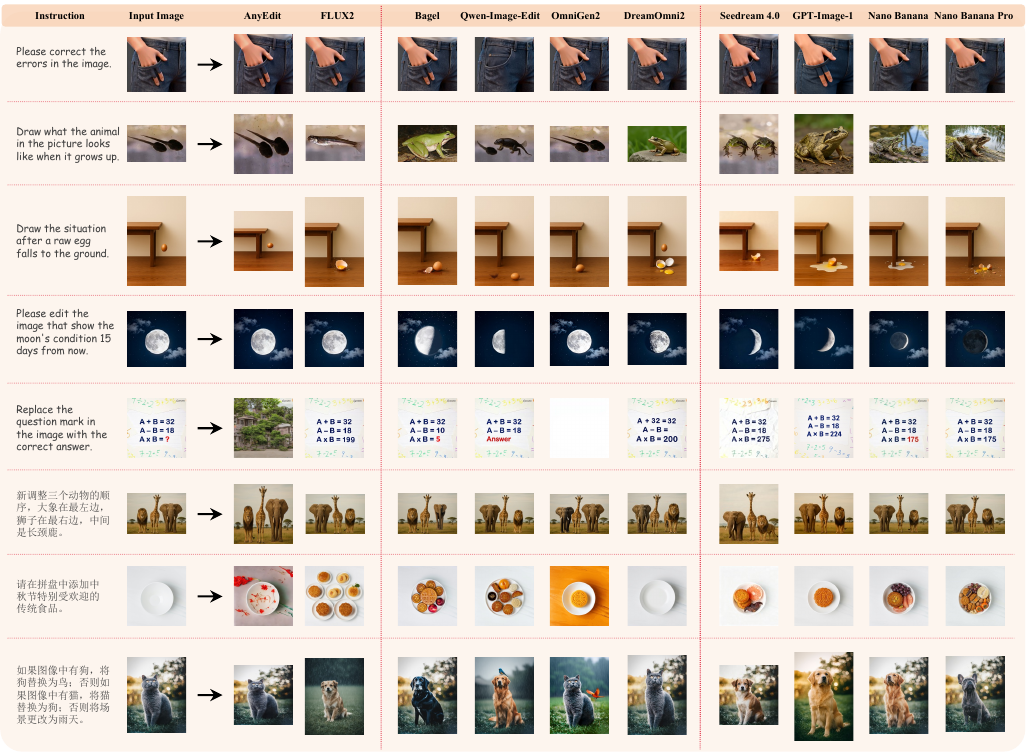}
\vspace{-1em}
\centering\caption{\label{fig:task2} More qualitative comparisons on the Interpretation task.}

\end{figure*}

\begin{figure*}[t]
\includegraphics[width=\linewidth]{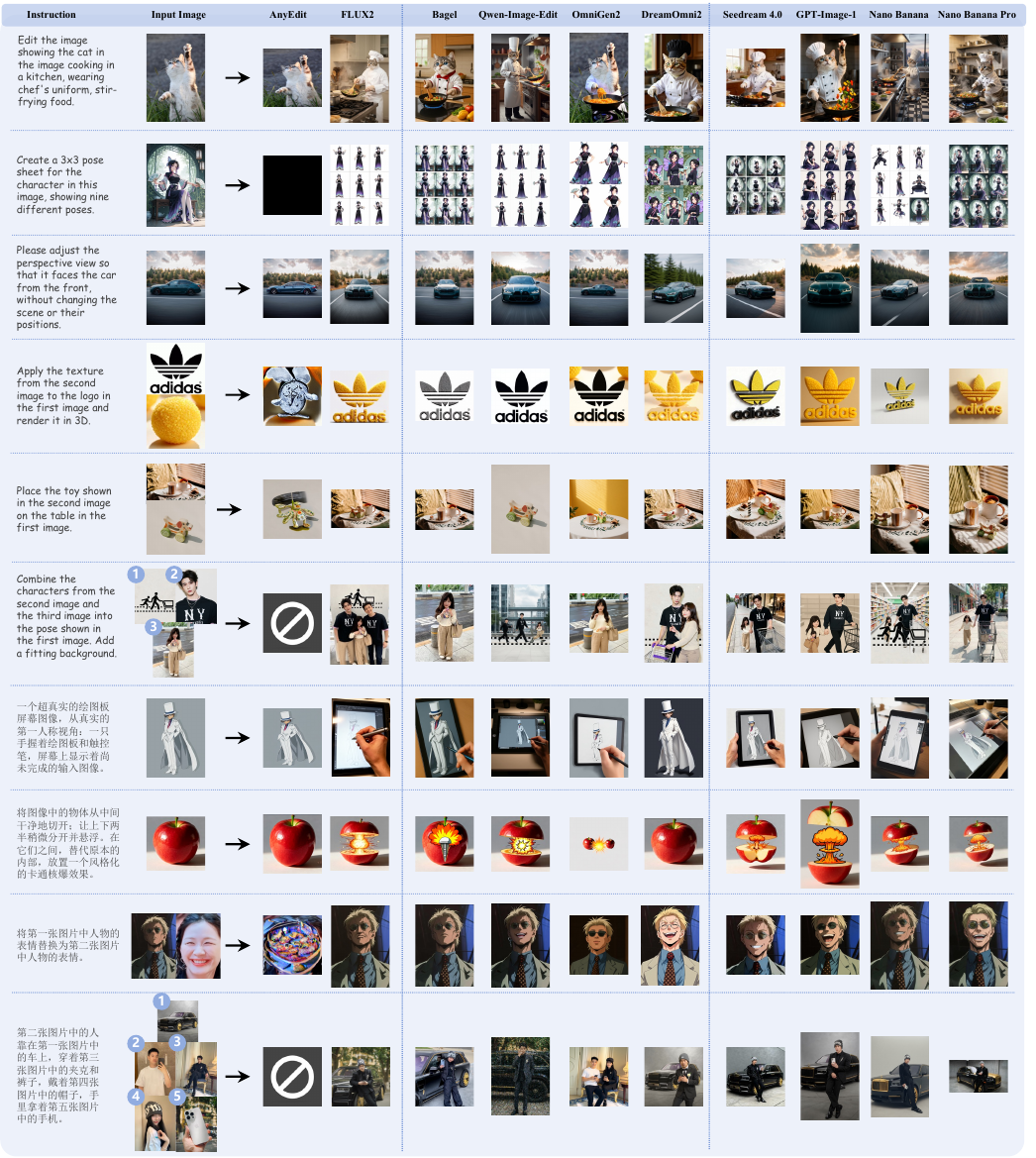}
\vspace{-1em}
\centering\caption{\label{fig:task3} More qualitative comparisons on the Imagination task.}

\end{figure*}

\begin{figure*}[t]
\includegraphics[width=\linewidth]{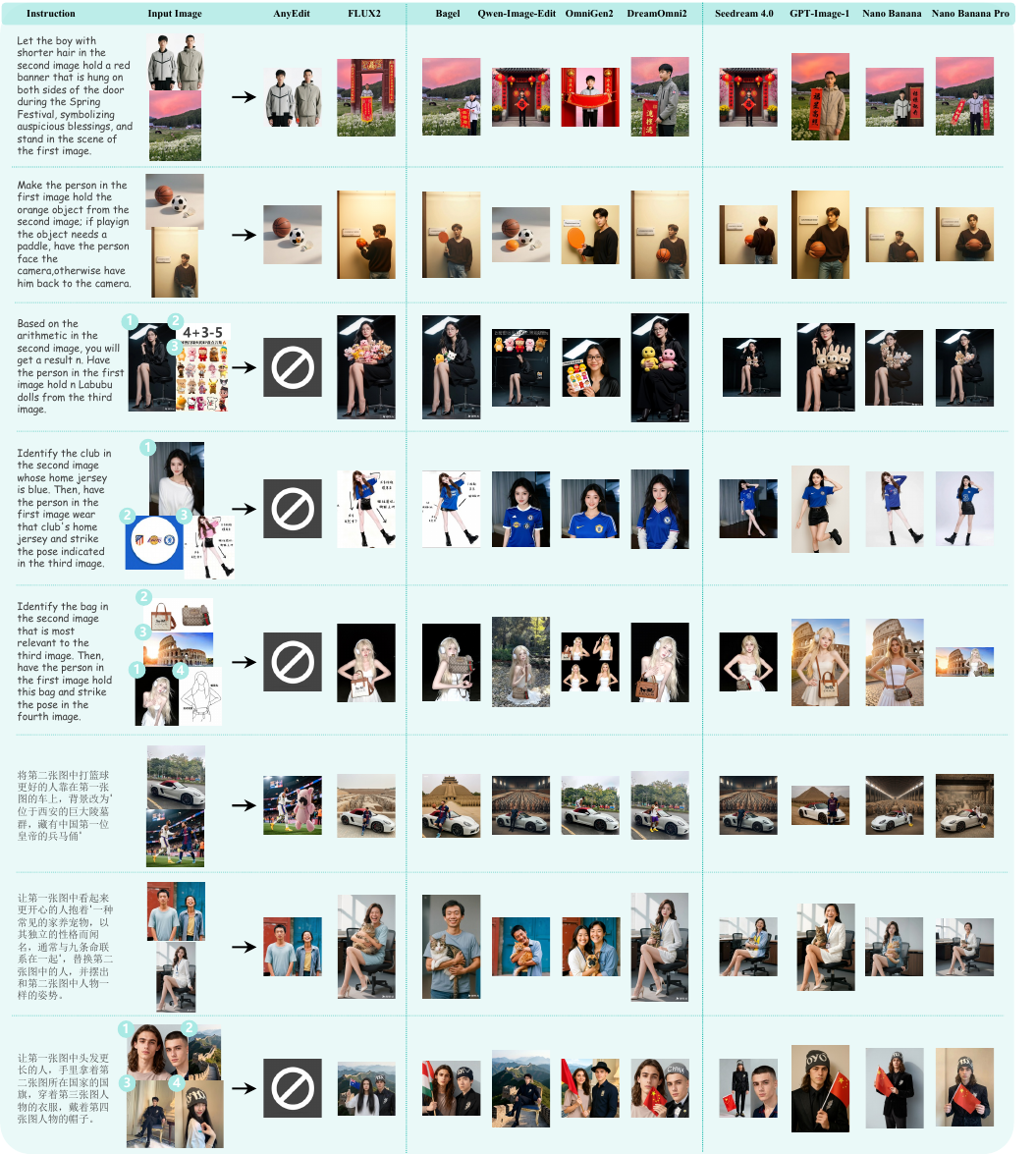}
\vspace{-1em}
\centering\caption{\label{fig:task4} More qualitative comparisons on the WiseEdit-Complex task.}

\end{figure*}

\paragraph{More Qualitative Comparisons. } In Fig.~\ref{fig:task1}, Fig.~\ref{fig:task2}, Fig.~\ref{fig:task3}, and Fig.~\ref{fig:task4}, we present additional qualitative comparisons across four tasks, \textit{i.e.}, Awareness, Interpretation, Imagination, and WiseEdit-Complex. 
We show the generated results of nine models, \textit{e.g.}, AnyEdit, FLUX.1 Kontext Dev, FLUX.2 Dev, Bagel, Qwen-Image-Edit, OmniGen2, DreamOmni2, Seedream 4.0, GPT-Image-1, Nano Banana and Nano Banana Pro. 
For any model that is unable to process a given input, its output is indicated by an image with a prohibition symbol.

\begin{table*}[t]
    \centering
    \caption{ \label{tab:single} Main results on WiseEdit across all cases or only single-image inputs on English version of instructions. For the majority of models, the average performance across all cases is  lower than the average performance observed for single-image input.}
    \vspace{-1em}
    \resizebox{1.0\linewidth}{!}{
        \begin{tabular}{lc|ccccc|ccccc|ccccc|c}
            \toprule
             \multirow{2}{*}{Model}    & \multirow{2}{*}{\textit{Cases}}&\multicolumn{5}{c}{\textbf{Awareness Task}}  &
            
                    \multicolumn{5}{c}{\textbf{Imagination Task}}  &
                    
                   \textbf{Overall} \\
            &  & $\mathcal{IF}\uparrow$ & $\mathcal{DP}\uparrow$ & $\mathcal{VQ}\uparrow$ & $\mathcal{KF}\uparrow$ & $\cellcolor[HTML]{E2E2E2}\mathbf{AVG}$ & $\mathcal{IF}\uparrow$ & $\mathcal{DP}\uparrow$ & $\mathcal{VQ}\uparrow$ & $\mathcal{CF}\uparrow$ & $\cellcolor[HTML]{E2E2E2}\mathbf{AVG}$& $\cellcolor[HTML]{C8C8C8}\mathbf{AVG}$\\ \hline
\multicolumn{13}{c}{\textit{English Version}} \\ \hline
\multirow{2}{*}{\textbf{InstructPix2Pix} }  & \textit{all cases} & 24.6 & 33.7 & 50.6 & 26.4 & \cellcolor[HTML]{E2E2E2}33.8 & 17.0 & 29.9 & 41.2 & 27.4 & \cellcolor[HTML]{E2E2E2}28.8 & \cellcolor[HTML]{C8C8C8}34.4 \\ 
& \textit{sing-img cases} &33.2 \textcolor{red}{(+8.6)} & 44.9 \textcolor{red}{(+11.2)} & 66.7 \textcolor{red}{(+16.1)} & 35.5 \textcolor{red}{(+9.1)} & \cellcolor[HTML]{E2E2E2}45.1 \textcolor{red}{(+11.3)} & 25.9 \textcolor{red}{(+9.0)} & 44.5 \textcolor{red}{(+14.7)} & 60.8 \textcolor{red}{(+19.6)} & 40.9 \textcolor{red}{(+13.6)} & \cellcolor[HTML]{E2E2E2}43.1 \textcolor{red}{(+14.2)} & \cellcolor[HTML]{C8C8C8}42.8 \textcolor{red}{(+8.5)} \\ \hline
\multirow{2}{*}{\textbf{MagicBrush} }  & \textit{all cases} & 27.2 & 43.4 & 53.3 & 27.1 & \cellcolor[HTML]{E2E2E2}37.8 & 18.0 & 36.9 & 44.8 & 22.3 & \cellcolor[HTML]{E2E2E2}30.5 & \cellcolor[HTML]{C8C8C8}35.5 \\ 
& \textit{sing-img cases} &36.6 \textcolor{red}{(+9.3)} & 57.4 \textcolor{red}{(+14.0)} & 70.2 \textcolor{red}{(+16.9)} & 36.4 \textcolor{red}{(+9.3)} & \cellcolor[HTML]{E2E2E2}50.1 \textcolor{red}{(+12.4)} & 27.5 \textcolor{red}{(+9.4)} & 54.6 \textcolor{red}{(+17.7)} & 66.1 \textcolor{red}{(+21.3)} & 33.7 \textcolor{red}{(+11.3)} & \cellcolor[HTML]{E2E2E2}45.5 \textcolor{red}{(+14.9)} & \cellcolor[HTML]{C8C8C8}44.6 \textcolor{red}{(+9.1)} \\ \hline
\multirow{2}{*}{\textbf{OmniGen} }  & \textit{all cases} & 35.0 & 42.0 & 46.7 & 37.4 & \cellcolor[HTML]{E2E2E2}40.3 & 42.2 & 35.1 & 46.0 & 38.7 & \cellcolor[HTML]{E2E2E2}40.5 & \cellcolor[HTML]{C8C8C8}36.6 \\ 
& \textit{sing-img cases} &34.6 \textcolor{blue}{(-0.4)} & 43.6 \textcolor{red}{(+1.6)} & 48.5 \textcolor{red}{(+1.9)} & 34.7 \textcolor{blue}{(-2.7)} & \cellcolor[HTML]{E2E2E2}40.3 \textcolor{red}{(+0.1)} & 47.9 \textcolor{red}{(+5.7)} & 35.7 \textcolor{red}{(+0.6)} & 44.7 \textcolor{blue}{(-1.2)} & 41.5 \textcolor{red}{(+2.8)} & \cellcolor[HTML]{E2E2E2}42.5 \textcolor{red}{(+2.0)} & \cellcolor[HTML]{C8C8C8}37.2 \textcolor{red}{(+0.7)} \\ \hline
\multirow{2}{*}{\textbf{AnyEdit} }  & \textit{all cases} & 25.0 & 54.6 & 61.3 & 26.3 & \cellcolor[HTML]{E2E2E2}41.8 & 9.1 & 49.7 & 50.9 & 16.5 & \cellcolor[HTML]{E2E2E2}31.5 & \cellcolor[HTML]{C8C8C8}37.7 \\ 
& \textit{sing-img cases} &32.9 \textcolor{red}{(+7.9)} & 68.0 \textcolor{red}{(+13.4)} & 72.8 \textcolor{red}{(+11.5)} & 33.5 \textcolor{red}{(+7.2)} & \cellcolor[HTML]{E2E2E2}51.8 \textcolor{red}{(+10.0)} & 12.9 \textcolor{red}{(+3.8)} & 70.1 \textcolor{red}{(+20.4)} & 66.6 \textcolor{red}{(+15.7)} & 18.4 \textcolor{red}{(+2.0)} & \cellcolor[HTML]{E2E2E2}42.0 \textcolor{red}{(+10.5)} & \cellcolor[HTML]{C8C8C8}44.5 \textcolor{red}{(+6.8)} \\ \hline
\multirow{2}{*}{\textbf{UltraEdit} }  & \textit{all cases} & 26.5 & 42.5 & 53.1 & 33.9 & \cellcolor[HTML]{E2E2E2}39.0 & 20.7 & 31.7 & 45.8 & 27.5 & \cellcolor[HTML]{E2E2E2}31.5 & \cellcolor[HTML]{C8C8C8}39.0 \\ 
& \textit{sing-img cases} &35.7 \textcolor{red}{(+9.1)} & 56.3 \textcolor{red}{(+13.8)} & 69.9 \textcolor{red}{(+16.8)} & 45.2 \textcolor{red}{(+11.3)} & \cellcolor[HTML]{E2E2E2}51.8 \textcolor{red}{(+12.8)} & 31.4 \textcolor{red}{(+10.6)} & 47.2 \textcolor{red}{(+15.5)} & 67.5 \textcolor{red}{(+21.7)} & 41.2 \textcolor{red}{(+13.6)} & \cellcolor[HTML]{E2E2E2}46.8 \textcolor{red}{(+15.4)} & \cellcolor[HTML]{C8C8C8}48.4 \textcolor{red}{(+9.4)} \\ \hline
\multirow{2}{*}{\textbf{ICEdit} }  & \textit{all cases} & 26.1 & 42.2 & 61.2 & 31.8 & \cellcolor[HTML]{E2E2E2}40.4 & 21.5 & 40.6 & 54.0 & 25.0 & \cellcolor[HTML]{E2E2E2}35.3 & \cellcolor[HTML]{C8C8C8}39.9 \\ 
& \textit{sing-img cases} &34.9 \textcolor{red}{(+8.8)} & 55.6 \textcolor{red}{(+13.4)} & 80.1 \textcolor{red}{(+18.9)} & 42.3 \textcolor{red}{(+10.4)} & \cellcolor[HTML]{E2E2E2}53.2 \textcolor{red}{(+12.9)} & 32.1 \textcolor{red}{(+10.6)} & 59.6 \textcolor{red}{(+19.0)} & 78.9 \textcolor{red}{(+24.9)} & 37.2 \textcolor{red}{(+12.1)} & \cellcolor[HTML]{E2E2E2}51.9 \textcolor{red}{(+16.6)} & \cellcolor[HTML]{C8C8C8}49.7 \textcolor{red}{(+9.8)} \\ \hline
\multirow{2}{*}{\textbf{FLUX.1 Kontext Dev} }  & \textit{all cases} & 31.4 & 52.0 & 55.0 & 35.5 & \cellcolor[HTML]{E2E2E2}43.5 & 39.1 & 47.1 & 43.4 & 27.1 & \cellcolor[HTML]{E2E2E2}39.2 & \cellcolor[HTML]{C8C8C8}43.2 \\ 
& \textit{sing-img cases} &40.7 \textcolor{red}{(+9.3)} & 67.2 \textcolor{red}{(+15.2)} & 71.0 \textcolor{red}{(+16.1)} & 45.9 \textcolor{red}{(+10.5)} & \cellcolor[HTML]{E2E2E2}56.2 \textcolor{red}{(+12.8)} & 55.9 \textcolor{red}{(+16.8)} & 67.5 \textcolor{red}{(+20.3)} & 62.1 \textcolor{red}{(+18.7)} & 38.6 \textcolor{red}{(+11.5)} & \cellcolor[HTML]{E2E2E2}56.0 \textcolor{red}{(+16.8)} & \cellcolor[HTML]{C8C8C8}53.1 \textcolor{red}{(+9.9)} \\ \hline

\multirow{2}{*}{\textbf{FLUX.2 Dev} } & \textit{all cases} & 42.6 & 63.3 & 78.4 & 53.3 & \cellcolor[HTML]{E2E2E2}59.4 & \color[HTML]{F88825}\textbf{73.6} & 70.7 & \color[HTML]{F88825}\textbf{82.1} & \color[HTML]{F88825}\textbf{43.6} & \cellcolor[HTML]{E2E2E2}\color[HTML]{F88825}\textbf{67.5} & \cellcolor[HTML]{C8C8C8}\color[HTML]{F88825}\textbf{61.8} \\ 

& \textit{sing-img cases} &44.5 \textcolor{red}{(+1.8)} & 62.6 \textcolor{blue}{(-0.8)} & 81.5 \textcolor{red}{(+3.1)} & 51.6 \textcolor{blue}{(-1.7)} & \cellcolor[HTML]{E2E2E2}60.0 \textcolor{red}{(+0.6)} & \color[HTML]{F88825}\textbf{78.8} \textcolor{red}{(+5.3)} & 68.9 \textcolor{blue}{(-1.8)} & 82.9 \textcolor{red}{(+0.8)} & \color[HTML]{F88825}\textbf{45.7} \textcolor{red}{(+2.1)} & \cellcolor[HTML]{E2E2E2}69.1 \textcolor{red}{(+1.6)} & \cellcolor[HTML]{C8C8C8}62.5 \textcolor{red}{(+0.7)} \\ \hline

\multirow{2}{*}{\textbf{Janus-4o} } & \textit{all cases} & 34.7 & 37.0 & 45.9 & 36.2 & \cellcolor[HTML]{E2E2E2}38.5 & 28.2 & 37.6 & 42.0 & 25.5 & \cellcolor[HTML]{E2E2E2}33.3 & \cellcolor[HTML]{C8C8C8}36.7 \\ 
& \textit{sing-img cases} &46.1 \textcolor{red}{(+11.5)} & 49.1 \textcolor{red}{(+12.2)} & 60.7 \textcolor{red}{(+14.7)} & 48.2 \textcolor{red}{(+11.9)} & \cellcolor[HTML]{E2E2E2}51.0 \textcolor{red}{(+12.6)} & 42.2 \textcolor{red}{(+13.9)} & 55.6 \textcolor{red}{(+18.0)} & 62.0 \textcolor{red}{(+20.0)} & 38.2 \textcolor{red}{(+12.7)} & \cellcolor[HTML]{E2E2E2}49.5 \textcolor{red}{(+16.2)} & \cellcolor[HTML]{C8C8C8}46.2 \textcolor{red}{(+9.6)} \\ \hline
\multirow{2}{*}{\textbf{UniWorld-V1} } & \textit{all cases} & 31.5 & 48.9 & 58.8 & 38.6 & \cellcolor[HTML]{E2E2E2}44.5 & 30.3 & 50.3 & 64.2 & 27.5 & \cellcolor[HTML]{E2E2E2}43.1 & \cellcolor[HTML]{C8C8C8}41.1 \\ 
& \textit{sing-img cases} &31.9 \textcolor{red}{(+0.5)} & 46.8 \textcolor{blue}{(-2.1)} & 59.5 \textcolor{red}{(+0.6)} & 39.1 \textcolor{red}{(+0.5)} & \cellcolor[HTML]{E2E2E2}44.3 \textcolor{blue}{(-0.1)} & 31.9 \textcolor{red}{(+1.6)} & 56.3 \textcolor{red}{(+5.9)} & 66.2 \textcolor{red}{(+2.0)} & 26.3 \textcolor{blue}{(-1.2)} & \cellcolor[HTML]{E2E2E2}45.2 \textcolor{red}{(+2.1)} & \cellcolor[HTML]{C8C8C8}41.8 \textcolor{red}{(+0.7)} \\ \hline
\multirow{2}{*}{\textbf{HiDream-E1} } & \textit{all cases} & 29.7 & 41.2 & 56.3 & 32.0 & \cellcolor[HTML]{E2E2E2}39.8 & 39.6 & 40.1 & 49.9 & 29.6 & \cellcolor[HTML]{E2E2E2}39.8 & \cellcolor[HTML]{C8C8C8}41.4 \\ 
& \textit{sing-img cases} &39.7 \textcolor{red}{(+10.1)} & 54.6 \textcolor{red}{(+13.4)} & 74.1 \textcolor{red}{(+17.7)} & 42.7 \textcolor{red}{(+10.7)} & \cellcolor[HTML]{E2E2E2}52.8 \textcolor{red}{(+13.0)} & 58.5 \textcolor{red}{(+18.9)} & 59.3 \textcolor{red}{(+19.2)} & 73.3 \textcolor{red}{(+23.5)} & 44.1 \textcolor{red}{(+14.5)} & \cellcolor[HTML]{E2E2E2}58.8 \textcolor{red}{(+19.0)} & \cellcolor[HTML]{C8C8C8}52.1 \textcolor{red}{(+10.7)} \\ \hline
\multirow{2}{*}{\textbf{OmniGen2} } & \textit{all cases} & 35.0 & 64.0 & 75.4 & 41.3 & \cellcolor[HTML]{E2E2E2}53.9 & 42.0 & 64.4 & 74.6 & 31.8 & \cellcolor[HTML]{E2E2E2}53.2 & \cellcolor[HTML]{C8C8C8}49.4 \\ 
& \textit{sing-img cases} &36.4 \textcolor{red}{(+1.4)} & 65.6 \textcolor{red}{(+1.6)} & 75.3 \textcolor{blue}{(-0.1)} & 41.2 \textcolor{blue}{(-0.1)} & \cellcolor[HTML]{E2E2E2}54.6 \textcolor{red}{(+0.7)} & 41.1 \textcolor{blue}{(-0.9)} & 70.9 \textcolor{red}{(+6.5)} & 74.7 \textcolor{red}{(+0.1)} & 29.6 \textcolor{blue}{(-2.3)} & \cellcolor[HTML]{E2E2E2}54.1 \textcolor{red}{(+0.9)} & \cellcolor[HTML]{C8C8C8}49.9 \textcolor{red}{(+0.5)} \\ \hline
\multirow{2}{*}{\textbf{Step1X-Edit-v1p2} } & \textit{all cases} & 39.8 & 53.5 & 61.3 & 44.4 & \cellcolor[HTML]{E2E2E2}49.7 & 44.7 & 49.4 & 50.3 & 28.4 & \cellcolor[HTML]{E2E2E2}43.2 & \cellcolor[HTML]{C8C8C8}49.5 \\ 
& \textit{sing-img cases} &52.9 \textcolor{red}{(+13.1)} & 70.5 \textcolor{red}{(+17.0)} & 80.6 \textcolor{red}{(+19.3)} & 58.8 \textcolor{red}{(+14.4)} & \cellcolor[HTML]{E2E2E2}65.7 \textcolor{red}{(+15.9)} & 66.0 \textcolor{red}{(+21.3)} & 72.8 \textcolor{red}{(+23.4)} & 74.1 \textcolor{red}{(+23.8)} & 42.5 \textcolor{red}{(+14.1)} & \cellcolor[HTML]{E2E2E2}63.9 \textcolor{red}{(+20.7)} & \cellcolor[HTML]{C8C8C8}61.7 \textcolor{red}{(+12.2)} \\ \hline
\multirow{2}{*}{\textbf{Echo-4o} } & \textit{all cases} & 47.6 & 63.0 & 75.4 & 51.7 & \cellcolor[HTML]{E2E2E2}59.4 & 63.4 & 62.4 & 73.7 & 41.2 & \cellcolor[HTML]{E2E2E2}60.2 & \cellcolor[HTML]{C8C8C8}57.8 \\ 
& \textit{sing-img cases} &52.7 \textcolor{red}{(+5.1)} & 65.8 \textcolor{red}{(+2.9)} & 77.7 \textcolor{red}{(+2.3)} & 53.7 \textcolor{red}{(+2.0)} & \cellcolor[HTML]{E2E2E2}62.5 \textcolor{red}{(+3.1)} & 70.8 \textcolor{red}{(+7.4)} & 65.2 \textcolor{red}{(+2.8)} & 72.7 \textcolor{blue}{(-1.0)} & 44.0 \textcolor{red}{(+2.8)} & \cellcolor[HTML]{E2E2E2}63.2 \textcolor{red}{(+3.0)} & \cellcolor[HTML]{C8C8C8}59.8 \textcolor{red}{(+2.0)} \\ \hline
\multirow{2}{*}{\textbf{Bagel} } & \textit{all cases} & 46.2 & 71.0 & 75.8 & 50.8 & \cellcolor[HTML]{E2E2E2}61.0 & 62.8 & \color[HTML]{F88825}\textbf{68.5} & 74.5 & 40.7 & \cellcolor[HTML]{E2E2E2}61.6 & \cellcolor[HTML]{C8C8C8}60.0 \\ 
& \textit{sing-img cases} &50.5 \textcolor{red}{(+4.3)} & 73.3 \textcolor{red}{(+2.3)} & 79.8 \textcolor{red}{(+3.9)} & 50.0 \textcolor{blue}{(-0.8)} & \cellcolor[HTML]{E2E2E2}63.4 \textcolor{red}{(+2.4)} & 72.4 \textcolor{red}{(+9.6)} & \color[HTML]{F88825}\textbf{75.0} \textcolor{red}{(+6.5)} & 75.1 \textcolor{red}{(+0.6)} & 42.8 \textcolor{red}{(+2.1)} & \cellcolor[HTML]{E2E2E2}66.3 \textcolor{red}{(+4.7)} & \cellcolor[HTML]{C8C8C8}62.3 \textcolor{red}{(+2.4)} \\ \hline

\multirow{2}{*}{\textbf{Uni-CoT} } & \textit{all cases} & 46.0 & 69.1 & 77.8 & 51.6 & \cellcolor[HTML]{E2E2E2}61.1 & 67.6 & 64.3 & 79.6 & 42.9 & \cellcolor[HTML]{E2E2E2}63.6 & \cellcolor[HTML]{C8C8C8}60.1 \\ 

& \textit{sing-img cases} &50.6 \textcolor{red}{(+4.7)} & 71.7 \textcolor{red}{(+2.6)} & 79.6 \textcolor{red}{(+1.7)} & 51.7 \textcolor{red}{(+0.2)} & \cellcolor[HTML]{E2E2E2}63.4 \textcolor{red}{(+2.3)} & 76.0 \textcolor{red}{(+8.4)} & 67.8 \textcolor{red}{(+3.6)} & 80.0 \textcolor{red}{(+0.5)} & 45.2 \textcolor{red}{(+2.3)} & \cellcolor[HTML]{E2E2E2}67.3 \textcolor{red}{(+3.7)} & \cellcolor[HTML]{C8C8C8}62.0 \textcolor{red}{(+2.0)} \\ \hline

\multirow{2}{*}{\textbf{Qwen-Image-Edit} } & \textit{all cases} & \color[HTML]{F88825}\textbf{48.1} & 69.0 & 79.5 & \color[HTML]{F88825}\textbf{53.6} & \cellcolor[HTML]{E2E2E2}62.5 & 67.1 & 66.8 & 79.2 & 42.3 & \cellcolor[HTML]{E2E2E2}63.8 & \cellcolor[HTML]{C8C8C8}60.2 \\ 

& \textit{sing-img cases} &\color[HTML]{F88825}\textbf{57.7} \textcolor{red}{(+9.6)} & 73.2 \textcolor{red}{(+4.2)} & 82.5 \textcolor{red}{(+3.0)} & \color[HTML]{F88825}\textbf{60.8} \textcolor{red}{(+7.2)} & \cellcolor[HTML]{E2E2E2}\color[HTML]{F88825}\textbf{68.5} \textcolor{red}{(+6.0)} & \color[HTML]{F88825}\textbf{78.8} \textcolor{red}{(+11.6)} & 73.7 \textcolor{red}{(+6.9)} & 81.9 \textcolor{red}{(+2.7)} & 45.1 \textcolor{red}{(+2.8)} & \cellcolor[HTML]{E2E2E2}\color[HTML]{F88825}\textbf{69.9} \textcolor{red}{(+6.0)} & \cellcolor[HTML]{C8C8C8}\color[HTML]{F88825}\textbf{64.2} \textcolor{red}{(+4.0)} \\ \hline

\multirow{2}{*}{\textbf{DreamOmni2} } & \textit{all cases} & 43.3 & \color[HTML]{F88825}\textbf{74.4} & \color[HTML]{F88825}\textbf{85.0} & 51.2 & \cellcolor[HTML]{E2E2E2}\color[HTML]{F88825}\textbf{63.5} & 50.6 & 64.9 & 81.9 & 35.3 & \cellcolor[HTML]{E2E2E2}58.2 & \cellcolor[HTML]{C8C8C8}60.6 \\ 
& \textit{sing-img cases} &46.7 \textcolor{red}{(+3.4)} & \color[HTML]{F88825}\textbf{77.1} \textcolor{red}{(+2.6)} & \color[HTML]{F88825}\textbf{86.0} \textcolor{red}{(+1.0)} & 52.5 \textcolor{red}{(+1.4)} & \cellcolor[HTML]{E2E2E2}65.6 \textcolor{red}{(+2.1)} & 54.1 \textcolor{red}{(+3.5)} & 68.0 \textcolor{red}{(+3.0)} & \color[HTML]{F88825}\textbf{84.2} \textcolor{red}{(+2.3)} & 34.5 \textcolor{blue}{(-0.8)} & \cellcolor[HTML]{E2E2E2}60.2 \textcolor{red}{(+2.0)} & \cellcolor[HTML]{C8C8C8}61.9 \textcolor{red}{(+1.4)} \\ \hline

\multirow{2}{*}{\textbf{Nano Banana} }  & \textit{all cases} & 70.6 & 85.7 & 86.8 & 75.2 & \cellcolor[HTML]{E2E2E2}79.6 & 75.3 & 73.8 & 87.3 & 44.3 & \cellcolor[HTML]{E2E2E2}70.2 & \cellcolor[HTML]{C8C8C8}75.0 \\ 
& \textit{sing-img cases} &69.6 \textcolor{blue}{(-0.9)} & 86.2 \textcolor{red}{(+0.5)} & 88.2 \textcolor{red}{(+1.4)} & 73.8 \textcolor{blue}{(-1.4)} & \cellcolor[HTML]{E2E2E2}79.5 \textcolor{blue}{(-0.1)} & 83.7 \textcolor{red}{(+8.4)} & 76.1 \textcolor{red}{(+2.3)} & 89.6 \textcolor{red}{(+2.3)} & 47.8 \textcolor{red}{(+3.5)} & \cellcolor[HTML]{E2E2E2}74.3 \textcolor{red}{(+4.1)} & \cellcolor[HTML]{C8C8C8}76.4 \textcolor{red}{(+1.3)} \\ \hline
\multirow{2}{*}{\textbf{Seedream 4.0} }  & \textit{all cases} & 70.8 & 78.1 & 86.6 & 74.6 & \cellcolor[HTML]{E2E2E2}77.5 & 82.2 & 77.8 & 86.9 & 47.0 & \cellcolor[HTML]{E2E2E2}73.5 & \cellcolor[HTML]{C8C8C8}75.2 \\ 

& \textit{sing-img cases} &70.2 \textcolor{blue}{(-0.5)} & 78.8 \textcolor{red}{(+0.7)} & 87.5 \textcolor{red}{(+0.9)} & 71.7 \textcolor{blue}{(-2.9)} & \cellcolor[HTML]{E2E2E2}77.1 \textcolor{blue}{(-0.5)} & 88.5 \textcolor{red}{(+6.2)} & \color[HTML]{319B62}\textbf{81.8} \textcolor{red}{(+3.9)} & 88.9 \textcolor{red}{(+2.0)} & 47.1 \textcolor{red}{(+0.1)} & \cellcolor[HTML]{E2E2E2}76.6 \textcolor{red}{(+3.1)} & \cellcolor[HTML]{C8C8C8}76.1 \textcolor{red}{(+0.9)} \\ \hline

\multirow{2}{*}{\textbf{GPT-image-1} }  & \textit{all cases} & 78.5 & 85.8 & \color[HTML]{319B62}\textbf{88.0} & 81.2 & \cellcolor[HTML]{E2E2E2}83.3 & 84.4 & 76.2 & \color[HTML]{319B62}\textbf{89.2} & 48.4 & \cellcolor[HTML]{E2E2E2}74.6 & \cellcolor[HTML]{C8C8C8}77.6 \\ 

& \textit{sing-img cases} &83.5 \textcolor{red}{(+5.0)} & 87.8 \textcolor{red}{(+2.0)} & \color[HTML]{319B62}\textbf{88.8} \textcolor{red}{(+0.8)} & 83.2 \textcolor{red}{(+2.0)} & \cellcolor[HTML]{E2E2E2}85.8 \textcolor{red}{(+2.5)} & \color[HTML]{319B62}\textbf{91.4} \textcolor{red}{(+7.1)} & 77.3 \textcolor{red}{(+1.1)} & \color[HTML]{319B62}\textbf{91.1} \textcolor{red}{(+1.9)} & 49.9 \textcolor{red}{(+1.5)} & \cellcolor[HTML]{E2E2E2}77.5 \textcolor{red}{(+2.9)} & \cellcolor[HTML]{C8C8C8}79.4 \textcolor{red}{(+1.8)} \\ \hline

\multirow{2}{*}{\textbf{Nano Banana Pro} }  & \textit{all cases} & \color[HTML]{319B62}\textbf{85.4} & \color[HTML]{319B62}\textbf{88.6} & 83.9 & \color[HTML]{319B62}\textbf{91.4} & \cellcolor[HTML]{E2E2E2}\color[HTML]{319B62}\textbf{87.3} & \color[HTML]{319B62}\textbf{86.6} & \color[HTML]{319B62}\textbf{79.5} & 88.8 & \color[HTML]{319B62}\textbf{51.5} & \cellcolor[HTML]{E2E2E2}\color[HTML]{319B62}\textbf{76.6} & \cellcolor[HTML]{C8C8C8}\color[HTML]{319B62}\textbf{82.4} \\ 

& \textit{sing-img cases} &\color[HTML]{319B62} \textbf{90.0} \textcolor{red}{(+4.6)} & \color[HTML]{319B62}\textbf{88.1} \textcolor{blue}{(-0.6)} & 84.3 \textcolor{red}{(+0.4)} & \color[HTML]{319B62}\textbf{91.2} \textcolor{blue}{(-0.2)} & \cellcolor[HTML]{E2E2E2}\color[HTML]{319B62}\textbf{88.4} \textcolor{red}{(+1.1)} & 89.7 \textcolor{red}{(+3.1)} & 78.9 \textcolor{blue}{(-0.7)} & 89.5 \textcolor{red}{(+0.7)} & \color[HTML]{319B62}\textbf{52.3} \textcolor{red}{(+0.8)} & \cellcolor[HTML]{E2E2E2} \color[HTML]{319B62}\textbf{77.6} \textcolor{red}{(+1.0)} & \cellcolor[HTML]{C8C8C8}\color[HTML]{319B62}\textbf{82.9} \textcolor{red}{(+0.5)} \\ 

            \bottomrule
        \end{tabular}}
    \vspace{-1.0em}
\end{table*}

\begin{table*}[t]
    \centering
    \caption{ \label{tab:single2} Main results on WiseEdit across all cases or only single-image inputs on Chinese version of instructions. For the majority of models, the average performance across all cases is  lower than the average performance observed for single-image input.}
    \vspace{-1em}
    \resizebox{1.0\linewidth}{!}{
        \begin{tabular}{lc|ccccc|ccccc|ccccc|c}
            \toprule
             \multirow{2}{*}{Model}    & \multirow{2}{*}{\textit{Cases}}&\multicolumn{5}{c}{\textbf{Awareness Task}}  &
            
                    \multicolumn{5}{c}{\textbf{Imagination Task}}  &
                    
                   \textbf{Overall} \\
            &  & $\mathcal{IF}\uparrow$ & $\mathcal{DP}\uparrow$ & $\mathcal{VQ}\uparrow$ & $\mathcal{KF}\uparrow$ & $\cellcolor[HTML]{E2E2E2}\mathbf{AVG}$ & $\mathcal{IF}\uparrow$ & $\mathcal{DP}\uparrow$ & $\mathcal{VQ}\uparrow$ & $\mathcal{CF}\uparrow$ & $\cellcolor[HTML]{E2E2E2}\mathbf{AVG}$& $\cellcolor[HTML]{C8C8C8}\mathbf{AVG}$\\ \hline

\multicolumn{13}{c}{\textit{Chinese Version}} \\ \hline
\multirow{2}{*}{\textbf{InstructPix2Pix} }  & \textit{all cases} & 14.2 & 51.0 & 58.6 & 18.1 & \cellcolor[HTML]{E2E2E2}35.5 & 3.7 & 52.0 & 53.2 & 9.0 & \cellcolor[HTML]{E2E2E2}29.5 & \cellcolor[HTML]{C8C8C8}34.1 \\ 
& \textit{sing-img cases} &19.8 \textcolor{red}{(+5.6)} & 67.2 \textcolor{red}{(+16.2)} & 77.0 \textcolor{red}{(+18.4)} & 24.8 \textcolor{red}{(+6.7)} & \cellcolor[HTML]{E2E2E2}47.2 \textcolor{red}{(+11.7)} & 6.9 \textcolor{red}{(+3.1)} & 76.4 \textcolor{red}{(+24.4)} & 78.2 \textcolor{red}{(+24.9)} & 14.4 \textcolor{red}{(+5.5)} & \cellcolor[HTML]{E2E2E2}44.0 \textcolor{red}{(+14.5)} & \cellcolor[HTML]{C8C8C8}42.9 \textcolor{red}{(+8.7)} \\ \hline
\multirow{2}{*}{\textbf{MagicBrush} }  & \textit{all cases} & 15.0 & 43.8 & 52.9 & 17.8 & \cellcolor[HTML]{E2E2E2}32.4 & 5.4 & 39.6 & 49.2 & 12.7 & \cellcolor[HTML]{E2E2E2}26.8 & \cellcolor[HTML]{C8C8C8}30.0 \\ 
& \textit{sing-img cases} &20.8 \textcolor{red}{(+5.8)} & 57.9 \textcolor{red}{(+14.1)} & 69.6 \textcolor{red}{(+16.7)} & 24.4 \textcolor{red}{(+6.6)} & \cellcolor[HTML]{E2E2E2}43.2 \textcolor{red}{(+10.8)} & 9.3 \textcolor{red}{(+3.9)} & 58.6 \textcolor{red}{(+19.0)} & 72.4 \textcolor{red}{(+23.2)} & 19.8 \textcolor{red}{(+7.1)} & \cellcolor[HTML]{E2E2E2}40.1 \textcolor{red}{(+13.3)} & \cellcolor[HTML]{C8C8C8}38.0 \textcolor{red}{(+8.0)} \\ \hline
\multirow{2}{*}{\textbf{OmniGen} }  & \textit{all cases} & 16.3 & 42.6 & 60.1 & 22.6 & \cellcolor[HTML]{E2E2E2}35.4 & 15.4 & 27.6 & 51.3 & 32.3 & \cellcolor[HTML]{E2E2E2}31.7 & \cellcolor[HTML]{C8C8C8}31.3 \\ 
& \textit{sing-img cases} &16.6 \textcolor{red}{(+0.3)} & 47.8 \textcolor{red}{(+5.1)} & 61.1 \textcolor{red}{(+1.0)} & 21.9 \textcolor{blue}{(-0.6)} & \cellcolor[HTML]{E2E2E2}36.8 \textcolor{red}{(+1.5)} & 16.8 \textcolor{red}{(+1.4)} & 32.5 \textcolor{red}{(+4.9)} & 52.3 \textcolor{red}{(+1.0)} & 33.2 \textcolor{red}{(+0.9)} & \cellcolor[HTML]{E2E2E2}33.7 \textcolor{red}{(+2.0)} & \cellcolor[HTML]{C8C8C8}32.5 \textcolor{red}{(+1.2)} \\ \hline
\multirow{2}{*}{\textbf{AnyEdit} }  & \textit{all cases} & 17.5 & 55.3 & 55.7 & 19.9 & \cellcolor[HTML]{E2E2E2}37.1 & 7.3 & 47.4 & 52.3 & 15.4 & \cellcolor[HTML]{E2E2E2}30.6 & \cellcolor[HTML]{C8C8C8}34.0 \\ 
& \textit{sing-img cases} &23.4 \textcolor{red}{(+5.9)} & 67.7 \textcolor{red}{(+12.4)} & 66.7 \textcolor{red}{(+11.0)} & 25.6 \textcolor{red}{(+5.7)} & \cellcolor[HTML]{E2E2E2}45.9 \textcolor{red}{(+8.7)} & 10.3 \textcolor{red}{(+3.0)} & 67.0 \textcolor{red}{(+19.5)} & 68.4 \textcolor{red}{(+16.1)} & 17.4 \textcolor{red}{(+2.0)} & \cellcolor[HTML]{E2E2E2}40.8 \textcolor{red}{(+10.2)} & \cellcolor[HTML]{C8C8C8}40.3 \textcolor{red}{(+6.3)} \\ \hline
\multirow{2}{*}{\textbf{UltraEdit} }  & \textit{all cases} & 16.9 & 58.5 & 62.1 & 21.2 & \cellcolor[HTML]{E2E2E2}39.7 & 9.3 & 42.3 & 51.8 & 14.8 & \cellcolor[HTML]{E2E2E2}29.5 & \cellcolor[HTML]{C8C8C8}38.9 \\ 
& \textit{sing-img cases} &23.3 \textcolor{red}{(+6.4)} & 76.9 \textcolor{red}{(+18.4)} & 81.5 \textcolor{red}{(+19.4)} & 28.8 \textcolor{red}{(+7.6)} & \cellcolor[HTML]{E2E2E2}52.6 \textcolor{red}{(+12.9)} & 14.8 \textcolor{red}{(+5.6)} & 62.5 \textcolor{red}{(+20.1)} & 76.2 \textcolor{red}{(+24.3)} & 22.8 \textcolor{red}{(+8.0)} & \cellcolor[HTML]{E2E2E2}44.1 \textcolor{red}{(+14.5)} & \cellcolor[HTML]{C8C8C8}48.0 \textcolor{red}{(+9.2)} \\ \hline
\multirow{2}{*}{\textbf{ICEdit} }  & \textit{all cases} & 12.9 & 29.1 & 63.6 & 17.4 & \cellcolor[HTML]{E2E2E2}30.8 & 5.1 & 37.4 & 56.5 & 16.7 & \cellcolor[HTML]{E2E2E2}28.9 & \cellcolor[HTML]{C8C8C8}32.7 \\ 
& \textit{sing-img cases} &18.1 \textcolor{red}{(+5.2)} & 39.0 \textcolor{red}{(+9.9)} & 83.5 \textcolor{red}{(+19.8)} & 23.9 \textcolor{red}{(+6.5)} & \cellcolor[HTML]{E2E2E2}41.1 \textcolor{red}{(+10.4)} & 8.9 \textcolor{red}{(+3.8)} & 55.4 \textcolor{red}{(+18.0)} & 82.9 \textcolor{red}{(+26.4)} & 25.6 \textcolor{red}{(+8.9)} & \cellcolor[HTML]{E2E2E2}43.2 \textcolor{red}{(+14.2)} & \cellcolor[HTML]{C8C8C8}40.9 \textcolor{red}{(+8.2)} \\ \hline
\multirow{2}{*}{\textbf{FLUX.1 Kontext Dev} }  & \textit{all cases} & 16.5 & 48.4 & 52.1 & 19.1 & \cellcolor[HTML]{E2E2E2}34.0 & 9.2 & 41.9 & 43.3 & 10.6 & \cellcolor[HTML]{E2E2E2}26.3 & \cellcolor[HTML]{C8C8C8}33.7 \\ 
& \textit{sing-img cases} &22.8 \textcolor{red}{(+6.3)} & 63.9 \textcolor{red}{(+15.5)} & 68.7 \textcolor{red}{(+16.5)} & 26.1 \textcolor{red}{(+7.0)} & \cellcolor[HTML]{E2E2E2}45.4 \textcolor{red}{(+11.3)} & 14.8 \textcolor{red}{(+5.6)} & 61.9 \textcolor{red}{(+20.0)} & 63.9 \textcolor{red}{(+20.6)} & 16.7 \textcolor{red}{(+6.2)} & \cellcolor[HTML]{E2E2E2}39.3 \textcolor{red}{(+13.1)} & \cellcolor[HTML]{C8C8C8}41.8 \textcolor{red}{(+8.1)} \\ \hline

\multirow{2}{*}{\textbf{FLUX.2 Dev} } & \textit{all cases} & 43.0 & 60.6 & 79.5 & 51.4 & \cellcolor[HTML]{E2E2E2}58.6 & \color[HTML]{F88825}\textbf{75.5} & \color[HTML]{F88825}\textbf{74.4} & 82.8 & \color[HTML]{F88825}\textbf{42.6} & \cellcolor[HTML]{E2E2E2}\color[HTML]{F88825}\textbf{68.8} & \cellcolor[HTML]{C8C8C8}\color[HTML]{F88825}\textbf{61.4} \\ 
& \textit{sing-img cases} &43.4 \textcolor{red}{(+0.4)} & 60.5 \textcolor{blue}{(-0.1)} & 83.6 \textcolor{red}{(+4.1)} & 47.8 \textcolor{blue}{(-3.5)} & \cellcolor[HTML]{E2E2E2}58.8 \textcolor{red}{(+0.2)} & \color[HTML]{F88825}\textbf{80.7} \textcolor{red}{(+5.2)} & \color[HTML]{F88825}\textbf{75.2} \textcolor{red}{(+0.8)} & 84.1 \textcolor{red}{(+1.3)} & 44.3 \textcolor{red}{(+1.8)} & \cellcolor[HTML]{E2E2E2}\color[HTML]{F88825}\textbf{71.1} \textcolor{red}{(+2.3)} & \cellcolor[HTML]{C8C8C8}62.2 \textcolor{red}{(+0.8)} \\ \hline
\multirow{2}{*}{\textbf{Janus-4o} } & \textit{all cases} & 31.1 & 38.6 & 46.3 & 34.1 & \cellcolor[HTML]{E2E2E2}37.5 & 25.4 & 36.9 & 41.8 & 22.1 & \cellcolor[HTML]{E2E2E2}31.5 & \cellcolor[HTML]{C8C8C8}35.6 \\ 
& \textit{sing-img cases} &41.6 \textcolor{red}{(+10.5)} & 51.3 \textcolor{red}{(+12.6)} & 61.1 \textcolor{red}{(+14.8)} & 45.4 \textcolor{red}{(+11.3)} & \cellcolor[HTML]{E2E2E2}49.8 \textcolor{red}{(+12.3)} & 38.0 \textcolor{red}{(+12.7)} & 54.6 \textcolor{red}{(+17.7)} & 61.7 \textcolor{red}{(+19.9)} & 33.3 \textcolor{red}{(+11.2)} & \cellcolor[HTML]{E2E2E2}46.9 \textcolor{red}{(+15.4)} & \cellcolor[HTML]{C8C8C8}44.9 \textcolor{red}{(+9.2)} \\ \hline
\multirow{2}{*}{\textbf{UniWorld-V1} } & \textit{all cases} & 18.4 & 49.0 & 60.0 & 26.2 & \cellcolor[HTML]{E2E2E2}38.4 & 17.9 & 54.8 & 68.9 & 18.0 & \cellcolor[HTML]{E2E2E2}39.9 & \cellcolor[HTML]{C8C8C8}37.7 \\ 
& \textit{sing-img cases} &18.9 \textcolor{red}{(+0.6)} & 49.1 \textcolor{red}{(+0.1)} & 61.7 \textcolor{red}{(+1.7)} & 25.3 \textcolor{blue}{(-0.9)} & \cellcolor[HTML]{E2E2E2}38.7 \textcolor{red}{(+0.4)} & 12.6 \textcolor{blue}{(-5.3)} & 62.2 \textcolor{red}{(+7.4)} & 71.3 \textcolor{red}{(+2.4)} & 15.5 \textcolor{blue}{(-2.5)} & \cellcolor[HTML]{E2E2E2}40.4 \textcolor{red}{(+0.5)} & \cellcolor[HTML]{C8C8C8}37.9 \textcolor{red}{(+0.3)} \\ \hline
\multirow{2}{*}{\textbf{HiDream-E1} } & \textit{all cases} & 28.2 & 37.6 & 51.4 & 32.4 & \cellcolor[HTML]{E2E2E2}37.4 & 32.5 & 39.2 & 47.5 & 27.4 & \cellcolor[HTML]{E2E2E2}36.6 & \cellcolor[HTML]{C8C8C8}38.5 \\ 
& \textit{sing-img cases} &37.9 \textcolor{red}{(+9.6)} & 49.9 \textcolor{red}{(+12.3)} & 67.7 \textcolor{red}{(+16.3)} & 43.3 \textcolor{red}{(+10.8)} & \cellcolor[HTML]{E2E2E2}49.7 \textcolor{red}{(+12.3)} & 48.3 \textcolor{red}{(+15.8)} & 58.0 \textcolor{red}{(+18.8)} & 69.9 \textcolor{red}{(+22.4)} & 40.9 \textcolor{red}{(+13.6)} & \cellcolor[HTML]{E2E2E2}54.3 \textcolor{red}{(+17.6)} & \cellcolor[HTML]{C8C8C8}48.5 \textcolor{red}{(+10.0)} \\ \hline
\multirow{2}{*}{\textbf{OmniGen2} } & \textit{all cases} & 35.1 & 57.9 & 72.4 & 41.0 & \cellcolor[HTML]{E2E2E2}51.6 & 45.5 & 64.0 & 72.0 & 33.8 & \cellcolor[HTML]{E2E2E2}53.8 & \cellcolor[HTML]{C8C8C8}48.8 \\ 
& \textit{sing-img cases} &38.9 \textcolor{red}{(+3.8)} & 58.9 \textcolor{red}{(+1.0)} & 73.0 \textcolor{red}{(+0.5)} & 42.7 \textcolor{red}{(+1.7)} & \cellcolor[HTML]{E2E2E2}53.4 \textcolor{red}{(+1.8)} & 42.8 \textcolor{blue}{(-2.7)} & 66.4 \textcolor{red}{(+2.5)} & 71.9 \textcolor{blue}{(-0.1)} & 31.2 \textcolor{blue}{(-2.6)} & \cellcolor[HTML]{E2E2E2}53.1 \textcolor{blue}{(-0.7)} & \cellcolor[HTML]{C8C8C8}49.1 \textcolor{red}{(+0.3)} \\ \hline
\multirow{2}{*}{\textbf{Step1X-Edit-v1p2} } & \textit{all cases} & 38.6 & 55.6 & 59.5 & 42.0 & \cellcolor[HTML]{E2E2E2}48.9 & 45.7 & 48.3 & 51.3 & 27.0 & \cellcolor[HTML]{E2E2E2}43.1 & \cellcolor[HTML]{C8C8C8}49.6 \\ 
& \textit{sing-img cases} &51.4 \textcolor{red}{(+12.7)} & 73.2 \textcolor{red}{(+17.6)} & 78.3 \textcolor{red}{(+18.8)} & 55.7 \textcolor{red}{(+13.7)} & \cellcolor[HTML]{E2E2E2}64.6 \textcolor{red}{(+15.7)} & 67.4 \textcolor{red}{(+21.8)} & 71.2 \textcolor{red}{(+22.9)} & 75.5 \textcolor{red}{(+24.2)} & 40.5 \textcolor{red}{(+13.5)} & \cellcolor[HTML]{E2E2E2}63.7 \textcolor{red}{(+20.6)} & \cellcolor[HTML]{C8C8C8}61.7 \textcolor{red}{(+12.1)} \\ \hline
\multirow{2}{*}{\textbf{Echo-4o} } & \textit{all cases} & 47.9 & 59.9 & 73.1 & \color[HTML]{F88825}\textbf{55.0} & \cellcolor[HTML]{E2E2E2}59.0 & 62.8 & 64.2 & 75.1 & 41.5 & \cellcolor[HTML]{E2E2E2}60.9 & \cellcolor[HTML]{C8C8C8}58.0 \\ 
& \textit{sing-img cases} &\color[HTML]{F88825}\textbf{53.3} \textcolor{red}{(+5.4)} & 62.8 \textcolor{red}{(+2.9)} & 74.7 \textcolor{red}{(+1.7)} & \color[HTML]{F88825}\textbf{57.2} \textcolor{red}{(+2.2)} & \cellcolor[HTML]{E2E2E2}62.0 \textcolor{red}{(+3.0)} & 70.3 \textcolor{red}{(+7.5)} & 64.4 \textcolor{red}{(+0.2)} & 74.6 \textcolor{blue}{(-0.5)} & 43.5 \textcolor{red}{(+2.1)} & \cellcolor[HTML]{E2E2E2}63.2 \textcolor{red}{(+2.3)} & \cellcolor[HTML]{C8C8C8}59.8 \textcolor{red}{(+1.8)} \\ \hline
\multirow{2}{*}{\textbf{Bagel} } & \textit{all cases} & \color[HTML]{F88825}\textbf{48.5} & 71.3 & 76.8 & 52.1 & \cellcolor[HTML]{E2E2E2}62.2 & 63.5 & 68.3 & 75.3 & 39.7 & \cellcolor[HTML]{E2E2E2}61.7 & \cellcolor[HTML]{C8C8C8}59.5 \\ 
& \textit{sing-img cases} &52.4 \textcolor{red}{(+3.9)} & 74.7 \textcolor{red}{(+3.4)} & 80.4 \textcolor{red}{(+3.6)} & 52.4 \textcolor{red}{(+0.3)} & \cellcolor[HTML]{E2E2E2}64.9 \textcolor{red}{(+2.8)} & 70.4 \textcolor{red}{(+6.9)} & 73.7 \textcolor{red}{(+5.3)} & 76.5 \textcolor{red}{(+1.1)} & 41.9 \textcolor{red}{(+2.2)} & \cellcolor[HTML]{E2E2E2}65.6 \textcolor{red}{(+3.9)} & \cellcolor[HTML]{C8C8C8}61.7 \textcolor{red}{(+2.2)} \\ \hline

\multirow{2}{*}{\textbf{Uni-CoT} } & \textit{all cases} & 46.2 & 70.0 & 80.7 & 53.6 & \cellcolor[HTML]{E2E2E2}\color[HTML]{F88825}\textbf{62.6} & 65.5 & 65.1 & 79.7 & 41.6 & \cellcolor[HTML]{E2E2E2}63.0 & \cellcolor[HTML]{C8C8C8}60.6 \\ 

& \textit{sing-img cases} &48.7 \textcolor{red}{(+2.5)} & 72.2 \textcolor{red}{(+2.1)} & 82.5 \textcolor{red}{(+1.8)} & 54.4 \textcolor{red}{(+0.8)} & \cellcolor[HTML]{E2E2E2}64.4 \textcolor{red}{(+1.8)} & 74.0 \textcolor{red}{(+8.5)} & 69.0 \textcolor{red}{(+3.9)} & 81.2 \textcolor{red}{(+1.4)} & 44.0 \textcolor{red}{(+2.4)} & \cellcolor[HTML]{E2E2E2}67.0 \textcolor{red}{(+4.1)} & \cellcolor[HTML]{C8C8C8}62.5 \textcolor{red}{(+2.0)} \\ \hline

\multirow{2}{*}{\textbf{Qwen-Image-Edit} } & \textit{all cases} & 45.0 & 67.3 & 79.9 & 52.9 & \cellcolor[HTML]{E2E2E2}61.3 & 66.3 & 67.2 & 80.0 & 41.7 & \cellcolor[HTML]{E2E2E2}63.8 & \cellcolor[HTML]{C8C8C8}60.6 \\ 

& \textit{sing-img cases} &51.7 \textcolor{red}{(+6.7)} & 70.8 \textcolor{red}{(+3.5)} & 84.1 \textcolor{red}{(+4.2)} & 57.0 \textcolor{red}{(+4.0)} & \cellcolor[HTML]{E2E2E2}\color[HTML]{F88825}\textbf{65.9} \textcolor{red}{(+4.6)} & 78.9 \textcolor{red}{(+12.6)} & 73.9 \textcolor{red}{(+6.7)} & 83.8 \textcolor{red}{(+3.7)} & \color[HTML]{F88825}\textbf{44.4} \textcolor{red}{(+2.7)} & \cellcolor[HTML]{E2E2E2}70.3 \textcolor{red}{(+6.4)} & \cellcolor[HTML]{C8C8C8}\color[HTML]{F88825}\textbf{64.3} \textcolor{red}{(+3.7)} \\ \hline

\multirow{2}{*}{\textbf{DreamOmni2} } & \textit{all cases} & 31.9 & \color[HTML]{F88825}\textbf{78.7} & \color[HTML]{F88825}\textbf{85.4} & 38.4 & \cellcolor[HTML]{E2E2E2}58.6 & 38.5 & 69.4 & \color[HTML]{F88825}\textbf{84.8} & 27.2 & \cellcolor[HTML]{E2E2E2}55.0 & \cellcolor[HTML]{C8C8C8} 56.0 \\ 

& \textit{sing-img cases} &34.2 \textcolor{red}{(+2.3)} & \color[HTML]{F88825}\textbf{84.1} \textcolor{red}{(+5.3)} & \color[HTML]{F88825}\textbf{88.5} \textcolor{red}{(+3.1)} & 37.7 \textcolor{blue}{(-0.7)} & \cellcolor[HTML]{E2E2E2}61.1 \textcolor{red}{(+2.5)} & 38.4 \textcolor{blue}{(-0.0)} & 74.7 \textcolor{red}{(+5.2)} & \color[HTML]{F88825}\textbf{87.0} \textcolor{red}{(+2.1)} & 25.5 \textcolor{blue}{(-1.6)} & \cellcolor[HTML]{E2E2E2}56.4 \textcolor{red}{(+1.4)} & \cellcolor[HTML]{C8C8C8}57.3 \textcolor{red}{(+1.3)} \\ \hline

\multirow{2}{*}{\textbf{Nano Banana} } & \textit{all cases} & 71.8 & 83.8 & 86.5 & 70.7 & \cellcolor[HTML]{E2E2E2}78.2 & 76.0 & 75.8 & 87.3 & 43.7 & \cellcolor[HTML]{E2E2E2}70.7 & \cellcolor[HTML]{C8C8C8}75.3 \\ & \textit{sing-img cases} &71.4 \textcolor{blue}{(-0.4)} & 84.3 \textcolor{red}{(+0.5)} & \color[HTML]{319B62}\textbf{88.2} \textcolor{red}{(+1.7)} & 67.9 \textcolor{blue}{(-2.8)} & \cellcolor[HTML]{E2E2E2}78.0 \textcolor{blue}{(-0.2)} & 84.9 \textcolor{red}{(+8.9)} & 77.7 \textcolor{red}{(+1.9)} & 90.5 \textcolor{red}{(+3.2)} & 46.8 \textcolor{red}{(+3.1)} & \cellcolor[HTML]{E2E2E2}75.0 \textcolor{red}{(+4.3)} & \cellcolor[HTML]{C8C8C8}76.6 \textcolor{red}{(+1.3)} \\ \hline \multirow{2}{*}{\textbf{Seedream 4.0} } & \textit{all cases} & 69.1 & 79.0 & 84.4 & 72.0 & \cellcolor[HTML]{E2E2E2}76.1 & 79.8 & \color[HTML]{319B62}\textbf{79.7} & 86.5 & 46.4 & \cellcolor[HTML]{E2E2E2}73.1 & \cellcolor[HTML]{C8C8C8}74.1 \\ & \textit{sing-img cases} &69.7 \textcolor{red}{(+0.6)} & 78.2 \textcolor{blue}{(-0.8)} & 85.0 \textcolor{red}{(+0.6)} & 69.4 \textcolor{blue}{(-2.6)} & \cellcolor[HTML]{E2E2E2}75.5 \textcolor{blue}{(-0.6)} & 87.5 \textcolor{red}{(+7.7)} & \color[HTML]{319B62}\textbf{81.6} \textcolor{red}{(+1.9)} & 88.8 \textcolor{red}{(+2.2)} & 46.6 \textcolor{red}{(+0.2)} & \cellcolor[HTML]{E2E2E2}76.1 \textcolor{red}{(+3.0)} & \cellcolor[HTML]{C8C8C8}74.9 \textcolor{red}{(+0.8)} \\ \hline \multirow{2}{*}{\textbf{GPT-image-1} } & \textit{all cases} & 77.0 & 80.7 & \color[HTML]{319B62}\textbf{86.6} & 80.7 & \cellcolor[HTML]{E2E2E2}81.2 & 78.8 & 73.3 & \color[HTML]{319B62}\textbf{89.6} & 48.3 & \cellcolor[HTML]{E2E2E2}72.5 & \cellcolor[HTML]{C8C8C8}76.2 \\ & \textit{sing-img cases} &83.0 \textcolor{red}{(+6.0)} & 80.6 \textcolor{blue}{(-0.2)} & 88.0 \textcolor{red}{(+1.4)} & 82.9 \textcolor{red}{(+2.3)} & \cellcolor[HTML]{E2E2E2}83.6 \textcolor{red}{(+2.4)} & 86.7 \textcolor{red}{(+7.9)} & 69.5 \textcolor{blue}{(-3.9)} & \color[HTML]{319B62}\textbf{91.4} \textcolor{red}{(+1.8)} & 50.0 \textcolor{red}{(+1.7)} & \cellcolor[HTML]{E2E2E2}74.4 \textcolor{red}{(+1.9)} & \cellcolor[HTML]{C8C8C8}77.6 \textcolor{red}{(+1.4)} \\ \hline \multirow{2}{*}{\textbf{Nano Banana Pro} } & \textit{all cases} & \color[HTML]{319B62}\textbf{84.6} & \color[HTML]{319B62}\textbf{91.8} & 83.1 & \color[HTML]{319B62}\textbf{87.9} & \cellcolor[HTML]{E2E2E2}\color[HTML]{319B62}\textbf{86.9} & \color[HTML]{319B62}\textbf{85.5} & 77.6 & 88.4 & \color[HTML]{319B62}\textbf{51.1} & \cellcolor[HTML]{E2E2E2}\color[HTML]{319B62}\textbf{75.6} & \cellcolor[HTML]{C8C8C8}\color[HTML]{319B62}\textbf{81.2} \\ 
& \textit{sing-img cases} &\color[HTML]{319B62}\textbf{86.5} \textcolor{red}{(+1.9)} & \color[HTML]{319B62}\textbf{92.5} \textcolor{red}{(+0.6)} & 83.3 \textcolor{red}{(+0.2)} & \color[HTML]{319B62}\textbf{87.6} \textcolor{blue}{(-0.2)} & \cellcolor[HTML]{E2E2E2}\color[HTML]{319B62}\textbf{87.5} \textcolor{red}{(+0.6)} & \color[HTML]{319B62}\textbf{91.0} \textcolor{red}{(+5.5)} & 77.4 \textcolor{blue}{(-0.2)} & 90.4 \textcolor{red}{(+2.1)} & \color[HTML]{319B62}\textbf{52.2} \textcolor{red}{(+1.1)} & \cellcolor[HTML]{E2E2E2}\color[HTML]{319B62}\textbf{77.8} \textcolor{red}{(+2.1)} & \cellcolor[HTML]{C8C8C8}\color[HTML]{319B62}\textbf{81.9} \textcolor{red}{(+0.7)} \\

            \bottomrule
        \end{tabular}}
    \vspace{-1.0em}
\end{table*}

\end{document}